%% file: ms.tex
\documentclass[letterpaper, 10 pt, conference]{ieeeconf}  

\IEEEoverridecommandlockouts                              

\usepackage[OT1]{fontenc} 

\usepackage{graphics} 
\usepackage{pgfplots}
\pgfplotsset{compat=newest}
\usepackage{subfig}

\usepackage{siunitx}
\usepackage{amsmath} 
\usepackage{amssymb} 

\usepackage{algorithm}
\usepackage[noend]{algpseudocode}

\makeatletter
\newcommand{\StatexIndent}[1][3]{%
	\setlength\@tempdima{\algorithmicindent}%
	\Statex\hskip\dimexpr#1\@tempdima\relax}
\algdef{S}[IF]{IfNoElse}[1]{\algorithmicif\ #1}%
\makeatother

\makeatletter
\newcommand\fs@betterruled{%
       \def\@fs@cfont{\bfseries}\let\@fs@capt\floatc@ruled
       \def\@fs@pre{\vspace*{5pt}\hrule height.8pt depth0pt \kern2pt}%
       \def\@fs@post{\kern2pt\hrule\relax}%
       \def\@fs@mid{\kern2pt\hrule\kern2pt}%
       \let\@fs@iftopcapt\iftrue}
\floatstyle{betterruled}
\restylefloat{algorithm}
\makeatother

\usepackage{booktabs} 
\usepackage{multirow} 

\usepackage{fancyhdr}
\newcommand{\reffig}[1]{Fig.~\ref{#1}}
\newcommand{\reftab}[1]{Tab.~\ref{#1}}
\newcommand{\refsec}[1]{Sec.~\ref{#1}}
\newcommand{\refeq}[1]{Eq.~\ref{#1}}
\newcommand{\refalg}[1]{Alg.~\ref{#1}}

\title{\LARGE \bf
Detection and Tracking of Small Objects\\ in Sparse 3D Laser Range Data
}

\author{Jan Razlaw, Jan Quenzel, and Sven Behnke
\thanks{All authors are with the Autonomous Intelligent Systems Group, University of Bonn, Germany
		{\tt\small \{razlaw, quenzel, behnke\}@ais.uni-bonn.de}}%
}

\begin{document}

\maketitle
\thispagestyle{fancy}
\pagestyle{empty}

\begin{abstract}
Detection and tracking of dynamic objects is a key feature for autonomous behavior in a continuously changing environment. 
With the increasing popularity and capability of micro aerial vehicles (MAVs) efficient algorithms have to be utilized to enable multi object tracking on limited hardware and data provided by lightweight sensors.
We present a novel segmentation approach based on a combination of median filters and an efficient pipeline for detection and tracking of small objects within sparse point clouds generated by a Velodyne VLP-16 sensor.
We achieve real-time performance on a single core of our MAV hardware by exploiting the inherent structure of the data. 
Our approach is evaluated on simulated and real scans of in- and outdoor environments, obtaining results comparable to the state of the art.
Additionally, we provide an application for filtering the dynamic and mapping the static part of the data, generating further insights into the performance of the pipeline on unlabeled data. 
\end{abstract}

\section{Introduction}

As robotics is getting more and more popular, autonomous robots are utilized in a growing variety of environments and situations.
One basis for the safe deployment of autonomous machines is a robust perception and anticipation of continuous changes in the world.
This problem is addressed by detection and tracking algorithms.
Detection consists of identifying or perceiving objects of interest, while tracking is the task of monitoring the objects' states over time. 
Knowledge about their temporal history allows to anticipate future behavior. 

Recent developments in the field of lightweight light detection and ranging (LiDAR) sensors facilitate their use on micro aerial vehicles (MAVs). 
These MAVs are utilized in an increasing number of applications, like mapping~\cite{droeschel2018efficient}, inventory~\cite{beul2018fast}, or even health care~\cite{kim2017drone}.
For those, collision avoidance and dynamic path planning ensure safety and enable the efficient usage of restricted resources with regards to energy consumption and flight time.
Detection and tracking of dynamic objects is a key feature for these tasks and interaction with the environment in general.

Another beneficiary of improvements in this field is autonomous driving.  
The cars are usually equipped with powerful computers and a variety of different sensors. 
MAVs, on the contrary, are constrained by their lifting capacity---hence, providing limited computational power and allowing lightweight sensors only.

\begin{figure}[t]
	\centering
	\includegraphics[width=\linewidth]{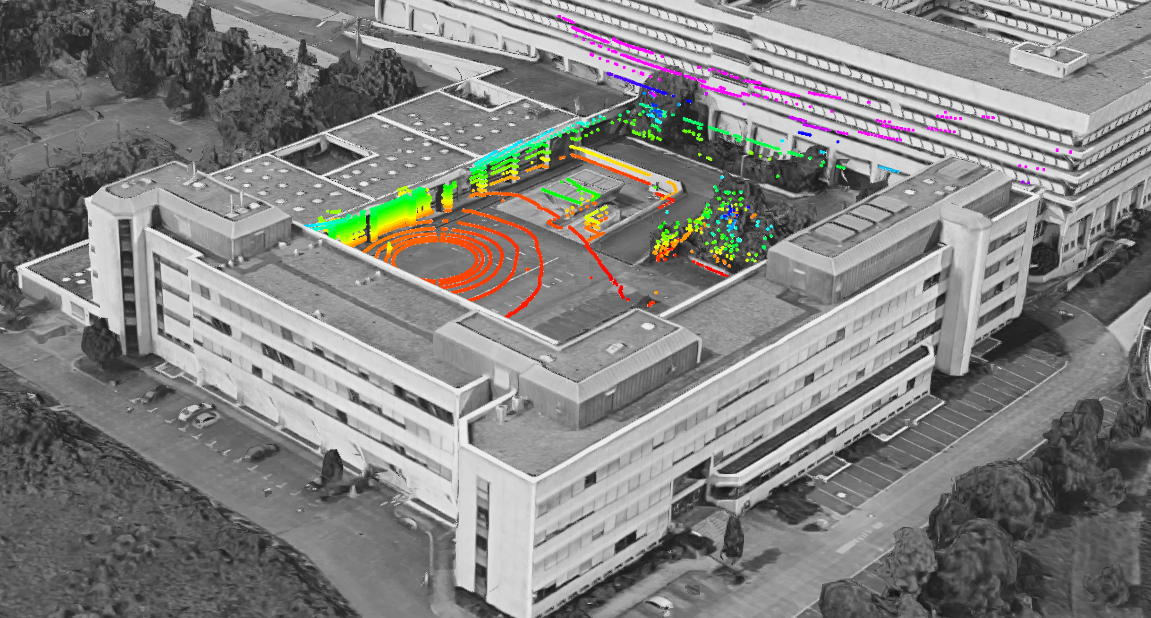}
	\caption[Exemplary Point Cloud]{An exemplary point cloud generated from one scan of the courtyard of the Landesbeh{\"o}rdenhaus in Bonn, projected into an aerial photography~\cite{lbh_maps}. Points are colored by height. }
	\label{fig:scan_by_height}
\end{figure}

Our goal is detecting and tracking multiple objects of a specific size---e.g. humans---in sparse point clouds as depicted in \reffig{fig:scan_by_height}.
These point clouds are generated by a Velodyne VLP-16 sensor mounted underneath a DJI Matrice 600 MAV.
Due to the limited compute power of the MAV, efficient algorithms have to be utilized for detection and tracking to achieve real-time performance. 

\section{Related Work}

Object tracking algorithms in general can be subdivided into two categories. 
In the first category are model-based tracking algorithms that utilize a detector to discriminate target objects from others based on a model description. 
The second category covers model-free tracking comparing consecutive scans of the environment to distinguish dynamic objects from the static background. 

The challenge for the former lies in the creation of a precise object model able to discriminate targets from non-targets while accounting for different settings and special cases.
Numerous works investigated the usage of trained classifiers~\cite{spinello2010layered, breiman2001random, inlida2017, farazionline} on a variety of features and descriptors.
Others utilized convolutional or recurrent neural networks~\cite{maturana20153d, ondruska2016deep, milan2017online} with promising results, but mostly on camera images.

Model-free tracking on the contrary is independent of a predefined model. 
Objects are detected either by searching for similar regions in consecutive scans implicitly building and updating a model~\cite{possegger2015defense} or, as usually applied for multi object tracking (MOT), by extracting the background and tracking the remaining measurement groups~\cite{moosmann2013joint, dewan2016motion}.
Such methods rely on the dynamics of objects as static or temporarily static objects are not tracked.

MOT algorithms additionally need to provide an assignment method capable of matching detections to corresponding tracked objects. 
Such assignment methods range from simple approaches minimizing pairwise distances of matches~\cite{inlida2017} to sophisticated but computationally more demanding joint probabilistic data association filters~\cite{schulz2003people} or likelihood-based sampling based methods~\cite{granstrom2018likelihood}.
Recent works~\cite{milan2017online, dequaire2017deep, farazionline} investigated the capability of recurrent neural networks, such as Long Short-Term Memory networks~\cite{hochreiter1997long}, for assignment and tracking. 

In this work, we attempt to combine the best of both worlds by relying on a simple, thus general, object model to not only preserve the ability of tracking static or temporary static targets but also to reduce parametrization effort and generalize to different settings. 
Additionally, we utilize the tracker's temporal information in the detector to reduce the rate of missed detections and concentrate on the usage of efficient algorithms to process data on limited hardware in real-time. 

\noindent In summary, the key features of our method are:
\begin{itemize}
\item A novel approach to segment point groups of a specified width range,
\item a detector utilizing segments, temporal information and the inherent structure of the data,
\item an efficient multi object tracker able to maintain tracks through short occlusions characteristic to the data,
\item real-time capability on a single CPU core of our MAV hardware,
\item and a practical application for filtering the dynamic and mapping the static part of the world.
\end{itemize}

\section{Method}

\begin{figure}[t]
    \vspace{0.2cm}
	\centering
	\includegraphics[width=0.8\linewidth]{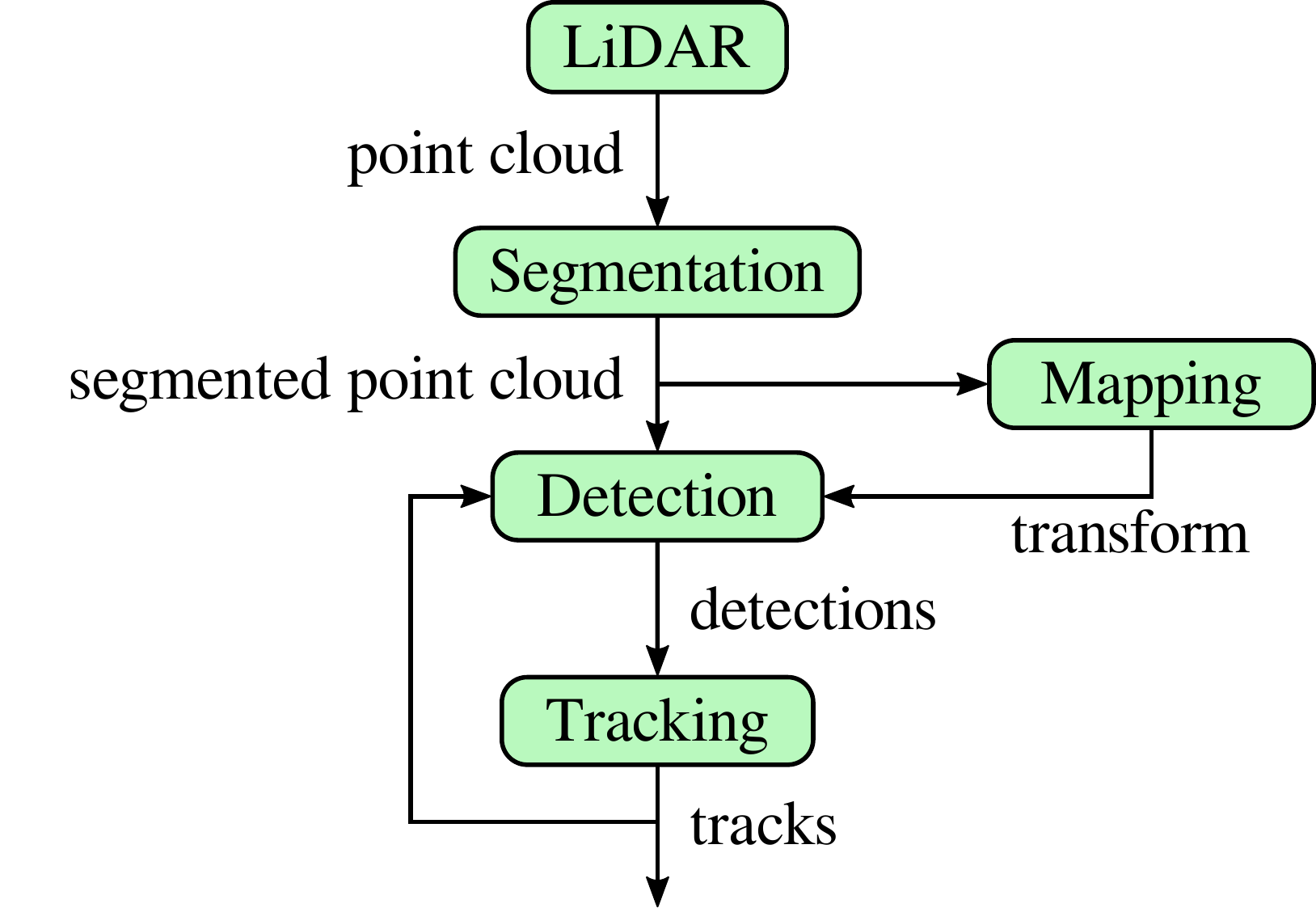}
	\caption[Overall Concept]{Overall concept of the MOT pipeline.}
	\label{fig:overall_concept}
\end{figure}

In the following, we provide a step-by-step description of the implemented MOT pipeline depicted in \reffig{fig:overall_concept}. 
Starting with the point cloud generated by the sensor, we preprocess the data by segmenting foreground point groups of a specified width range. 
This segmented cloud is used to create object detections that are fed to the MOT algorithm. 
The estimated tracks are then returned to the detector aiding the detection in following scans. 
We estimate the object states in the world frame. 
For this purpose, we utilize Multi Resolution Surfel Mapping~\cite{droeschel2014local} to estimate the sensors position in the world.
This mapping algorithm was explicitly developed to work in real-time on sparse laser range data.     

\subsection{Point Cloud Generation}\label{section:point_cloud_generation}

The chosen sensor to detect small objects in a long range is the Velodyne VLP-16.
The sensor has 16 laser-detector pairs placed on a vertical axis and oriented with an angle of $2^{\circ}$ to each other resulting in a $30^{\circ}$ vertical field of view (FoV).
This setup allows getting 16 measurements at a time. 
Spinning the laser-detector pairs around the sensor's vertical axis generates a $360^{\circ}$ horizontal scan of the environment consisting of 16 \textit{scan rings}.
\reffig{fig:scan_by_height} shows an example scan.

We exploit the configuration of $16 \times n$ measurements to generate one organized point cloud per full rotation, preserving the grid-like structure of the scan. 

\subsection{Segmentation}\label{sec:segmentation}

The scan rings generated by the sensor are deformed by objects in the environment, resulting in grouped measurements closer to the sensor than their neighboring measurements from the background.
Due to the sensors low vertical resolution, especially small or distant objects raise the risk of laying in between scan rings or corresponding to only very few measurements. 
Training of sophisticated object models under these circumstances is hard if not impossible. 
Hence, we segment objects according to their width, as this is the most distinct feature we can compute even for distant targets. 
Our goal is to find all points belonging to foreground groups of a specified width range.

We utilize the segmentation method we presented in~\cite{schwarzteam} consisting of two median filters---one for noise, one for background---with different kernel sizes applied to the distance readings of single scan rings. 
This method segments foreground point groups of a specified width, by filtering all narrower objects utilizing the \textit{noise filter} with a smaller kernel size and additionally filtering the target objects themselves using a slightly bigger kernel size in the \textit{background filter}.
Points for which the filters return different results are classified as target points. 
This approach naturally extends to segmenting groups of a specified width range by defining the minimal and maximal width explicitly through the noise and background filter kernel sizes. 

By exploiting the organized structure of the data, we introduce points with invalid distance readings---e.g. measurements in the direction of the sky or on absorbing surfaces. 
We automatically classify these as background points and replace the invalid distance by a fixed value exceeding the maximal measuring range. 
This way, invalid points can still be utilized by neighboring valid measurements during the median computation and allow segmenting objects in the sky, e.g. other MAVs.

\subsection{Detection}\label{sec:detection}

Detection generation is split up into clustering of segmented points and subsequent filtering.
For clustering, we utilize region growing on the organized structure of the point cloud. 
The region growing algorithm connects a seed point to its neighboring segment points and those successively to their neighboring segment points. 
The neighborhood is defined on the grid structure of the cloud.
We apply \refalg{alg:region_growing} to each unclustered segment point.

\begin{algorithm}[t]
	\caption{Region Growing Clustering (RGC)}\label{alg:region_growing}
	\begin{algorithmic}[1]
		\Function{RGC}{Organized grid of segmented points $P$}
		
		\State $C$ : empty list of clusters
		\State $Q$ : empty queue of points to process
		\State $c_{min}$ : minimal number of points within valid cluster
		\For{\textbf{each} not visited segment point $p_i \in P$}
		\State $c$ : empty cluster 
		\State add $p_i$ to $Q$ and $c$
		\State mark $p_i$ as visited 
		\While{$Q$ not empty}
		\State dequeue $p_j$ from $Q$
		\For{\textbf{each} neighbor $p_n$ : $\lVert p_n-p_j \rVert_{1} \le r$}
		\IfNoElse{$p_n$ not visited and a segment point and}
		\StatexIndent[4] \;\, $\lVert p_n-p_j \rVert_{2} < \theta$ \algorithmicthen
		\State add $p_n$ to $Q$ and $c$
		\State mark $p_n$ as visited
		\EndIf
		\EndFor
		\EndWhile
		\If{$|c| \geq c_{min}$} 
		\State add $c$ to $C$
		\EndIf
		\EndFor
		\State\Return $C$
		\EndFunction
	\end{algorithmic}
\end{algorithm}

\begin{figure}[t]
	\centering
	\includegraphics[width=\linewidth]{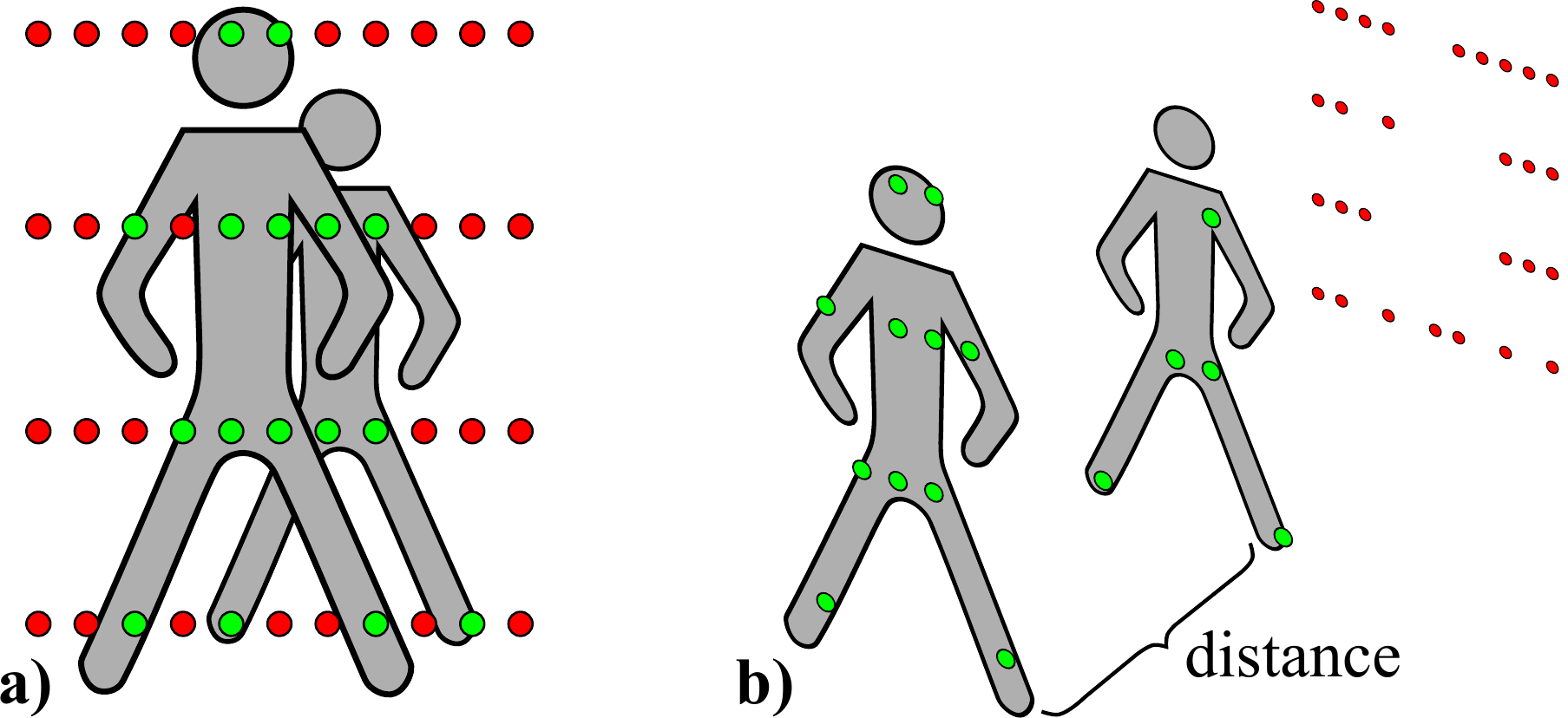}
	\caption[Region Growing Parameters]{Two persons scanned behind each other with green segment points and red background points. \textbf{a)} Sensor perspective: Simply connecting the direct neighbors would result in wrong clusters. \textbf{b)} Shifted perspective: Accounting for the distance between neighboring points helps to distinguish segment points corresponding to different objects.}
	\label{fig:region_growing_sensor_perspective}
\end{figure}

We adapt the clustering at two points to work more robustly on the special structure of our data. 
Due to the sparsity of the data, the neighborhood search radius in line 11 is extended to check more than just the direct neighbors on the grid (\reffig{fig:region_growing_sensor_perspective}a). 
Additionally, in line 12 the Euclidean distance of the current point to its neighbors has to be taken into account to prevent under-clustering of several distinct but partially occluding objects (\reffig{fig:region_growing_sensor_perspective}b). 

One drawback of the organized grid for this clustering method is a potential overlap of the start and end of a circular scan.  
We need to handle this case explicitly by computing the approximate overlap, finding the clusters lying within and fusing those corresponding to the same objects in the world. 

After clustering the segment points, we need to filter those clusters not matching our simple object model consisting of a height range and a maximal diagonal width.
Due to the sensor's limited vertical FoV, objects might be scanned partially. 
Consequently, we refrain from testing for a minimal height for clusters containing at least one point from the top and bottom scan rings.  
Additionally, we exploit the tracker's temporal information by relaxing the object model---omitting the minimal height and increasing the maximal width---for clusters in the vicinity of already tracked objects.  
Clusters fitting this description are considered valid detections. 

\subsection{Multi Object Tracking}\label{sec:tracking}

Multi Object Tracking in general is the task of monitoring the states of several objects simultaneously.
For this purpose, the algorithm maintains a set of object hypotheses.
These are represented by an axis aligned bounding box and a state consisting of a 3D position and velocity.
The hypotheses are updated using points corresponding to valid detections in the most current scan.
One Kalman filter with a constant velocity model is deployed for each hypothesis.

For tracking multiple objects, it is essential to know which detection corresponds to which object hypothesis. 
We solve this classical assignment problem in polynomial time utilizing the Hungarian method~\cite{munkres1957algorithms}.
The algorithm finds a one-to-one assignment for a given cost matrix minimizing the total assignment costs.
Hence, we model our problem as an adjacency matrix between detections and hypotheses utilizing the Bhattacharyya distance as a proximity measure. 
We forbid individual assignments that exceed a distance threshold.

Each assigned detection is used to correct the matched hypothesis state. 
We prevent velocities induced by sensor noise and varying amounts of partial occlusions by truncating velocity estimates of up to $1\,$km/h to zero. 
Additionally, we truncate estimated velocities to a maximum of $10\,$km/h to reduce the effect of volatile detections, due to sparse measurements on altering object parts. 

A detection without a matching hypothesis might be correspondent to an object entering the FoV and thus creates a new hypothesis. 
Detections generated using the relaxed object model implicitly correspond to already tracked objects---hence do not create new hypotheses. 
Unassigned hypotheses might correspond to objects that left the FoV and thus need to be tested for validity.
As unassigned hypotheses' covariances grow with each prediction step, we declare those with a high covariance as non-valid and initiate deletion.
For this, we compute the eigenvalues of the Kalman filter's covariance matrix and test if at least one value exceeds a threshold.  
To account for oversegmentation or misassignments, we delete the younger hypotheses being in the vicinity of older hypotheses. 

Additionally, we classify tracked objects into the classes \textit{static} and \textit{dynamic}. 
Each hypothesis is generated static and becomes dynamic once its current bounding box does not intersect with its initial bounding box.  

\subsection{Dynamic Objects Filter}\label{sec:object_filter}

As a practical application, we filter out tracked dynamic objects from all point clouds and generate a map of the static part of the scene. 
For this, we discard every detected point corresponding to a dynamic hypothesis from each currently processed point cloud. 
These filtered clouds are stored in a log.
Once a hypothesis becomes dynamic, all its previous bounding boxes---from the time it was static---are utilized to remove corresponding points from these logged clouds. 
If a dynamic hypothesis loses track of an object but recovers, we filter points corresponding to the predicted bounding boxes from intermediate steps similarly.
The result is a log of point clouds corresponding to the static part of the world.

\section{Evaluation}

We evaluate efficacy and efficiency of our proposed MOT approach as follows. 
We start by giving an overview of two metrics commonly used to evaluate MOT algorithms against ground truth annotations. 
These annotations are provided in two data sets we inspect afterwards.
We utilize both for evaluation and parameter optimization and discuss our achieved results.   

\subsection{Evaluation Metrics}

The first metric is the CLEAR MOT metric~\cite{bernardin2008evaluating}.
It utilizes the Hungarian method to assign hypotheses to ground truth labels in each time step. 
Valid matches have a Euclidean distance below a threshold---\SI{0.5}{\metre} are suggested by the authors for visual people tracking.
Matched hypothesis-label pairs from the previous time step are not reassigned to others if both are still present and close to each other. 
Using the sum of distances $d_t$ between each hypothesis and matched label for time $t$ and the number of matches $c_t$ it computes the Multi Object Tracking Precision (MOTP) as
\begin{align}\label{eq:mot_motp}
MOTP = \frac{\sum_{t}d_t}{\sum_{t}c_t}.
\end{align}
Additionally, it counts the number of missed labels $m_t$, false positives $fp_t$, mismatch errors $mme_t$ and labels $g_t$ for each time $t$ to compute the Multi Object Tracking Accuracy (MOTA) as 
\begin{align}\label{eq:mot_mota}
MOTA = 1 - \frac{\sum_{t}(m_t+fp_t+mme_t)}{\sum_{t}g_t}.
\end{align}
The MOTA is a measure for the consistency of the generated tracks.

Another way to evaluate the performance of an MOT algorithm is to inspect how much of the object tracks were covered by the hypotheses~\cite{li2009learning}. 
This metric is split up into three ratios: Mostly Tracked (MT), Partially Tracked (PT), and Mostly Lost (ML).
An object trajectory is mostly tracked if at least $80\%$ of it is covered by hypotheses. 
It is mostly lost if less than $20\%$ is covered and partially tracked for the remaining cases.   
We apply the same constraints as for the CLEAR MOT for an object to be classified as tracked.

This metric does not account for identity switches, false positives or precision.
It can be seen as an addition to the CLEAR MOT metric, providing a more detailed insight to the ratio of misses. 

\subsection{Data Sets}

Three data sets were utilized for evaluation and parameter optimization.
The first two are used for a quantitative evaluation against ground truth data. 
The third data set consists of scans recorded from our MAV setup during two flights in a large courtyard. 
Due to missing ground truth for the latter, we utilize our filtering application for a qualitative evaluation. 

\subsubsection{InLiDa}

The Indoor LiDAR Dataset \textit{InLiDa}~\cite{inlida2017} consists of six hand labeled sequences captured using a Velodyne VLP-16 in an indoor environment.
The sequences have a total duration of 501 seconds and contain 4823 scans.
The sensor is placed in a fixed location in a corridor or a hall. 
Up to eight dynamic objects---seven humans and one robot---are simultaneously visible and labeled at point-level. 
The data set provides challenging situations with occlusions, groups of close objects moving in the same direction, rapid velocity changes and other dynamic objects, like doors.  

\subsubsection{Simulated}

Additionally, we simulated the Velodyne VLP-16 in a set of diverse outdoor settings to generate another data set (\reffig{fig:datasets}).
For simulation, we used Gazebo~\cite{koenig2004design} and adapted a Velodyne VLP-16 simulator~\cite{velodyne_simulator} to generate organized point clouds. 
The data set provides sequences with a varying number of dynamic persons within static environments with different amounts of clutter and distractions (\reftab{tab:simulated_data}). 
Our simulated persons avoid obstacles, change their velocities from $3.5\,$km/h to $12.5\,$km/h, and pause from time to time.
The environments are limited to a distance of up to \SI{140}{\metre} to the sensor in the center at a height of \SI{2.5}{\metre}.
This implies that measurements of target objects do get very sparse or disappear completely due to occlusions or objects leaving the measurement range of the sensor.  
We recorded a data set using a static sensor and another set of sequences with a dynamic sensor moving on a trajectory with the shape of an 8 within a radius of about \SI{4}{\metre} around the center of the environment.    
Ground truth labels are provided at point-level. 

\begin{table}[t]
    \vspace{0.2cm}
	\setlength{\tabcolsep}{2pt}
	\begin{center}
		\caption{Properties of simulated sequences.}
		\begin{tabular}{c|ccccl}
			ID & Area				 & Targets & Duration & Scans & Environment \\ 
			\midrule
			1 			& 100m $\times$ 100m & 6 	& 98s &	986 & Empty field \\ 
			2 			& 100m $\times$ 100m & 50 	& 99s &	995 & Empty field \\ 
			3 			& 200m $\times$ 200m & 6 	& 97s &	978 & Empty field \\ 
			4 			& 100m $\times$ 100m & 6 	& 54s &	547 & Industrial \\ 
			5 			& 200m $\times$ 200m & 6 	& 92s &	925 & Industrial, Shops, Houses \\ 
			6 			& 100m $\times$ 150m & 6 	& 97s &	974 & Park
			
		\end{tabular}
		\label{tab:simulated_data}
	\end{center}
\end{table} 

\newcommand{\simWidth}{0.198\linewidth}
\begin{figure*}[t]
	\subfloat{
		\centering
		\includegraphics[width=0.17\linewidth]{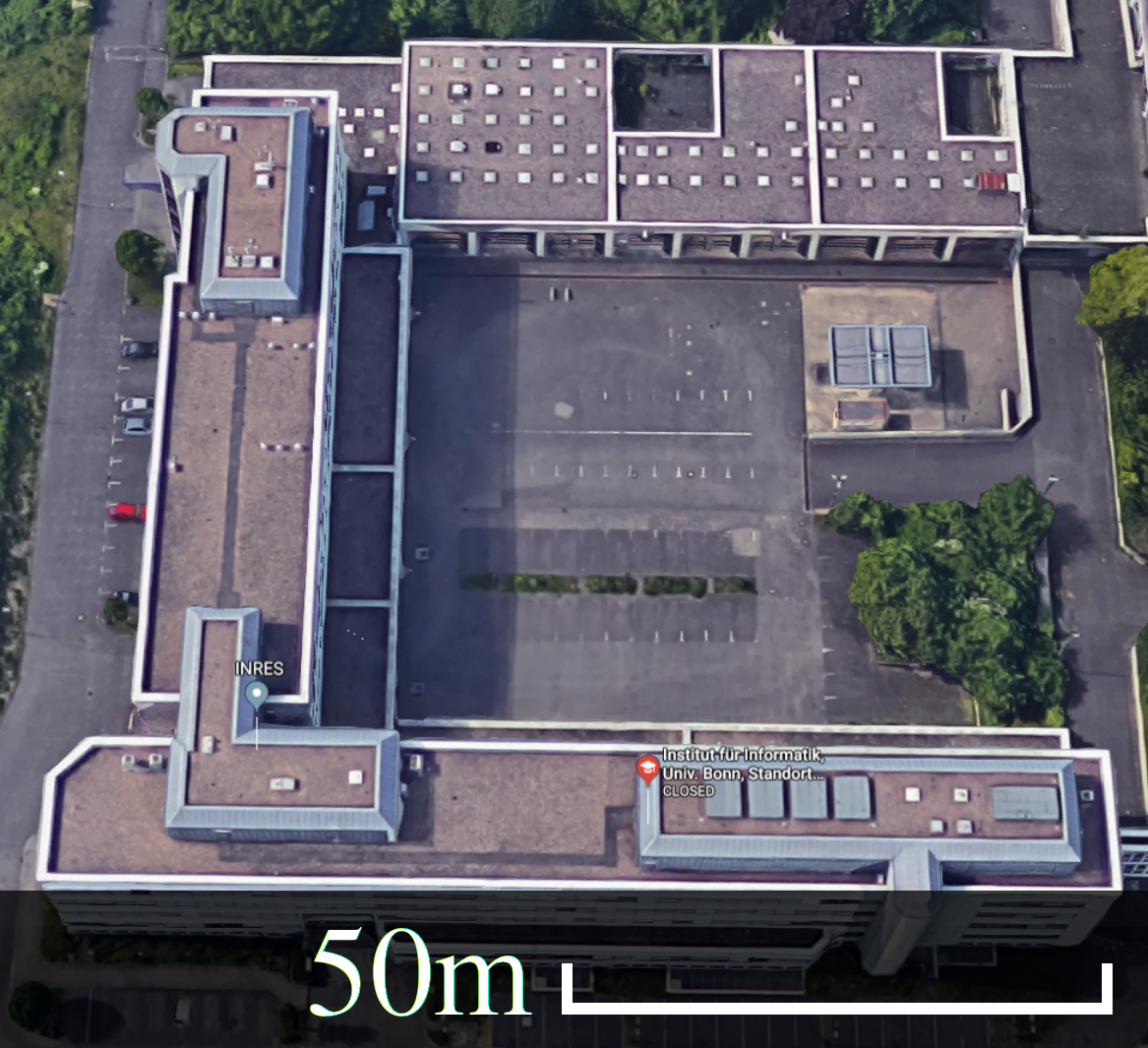}%
	}
	\subfloat{
		\centering
		\includegraphics[width=\simWidth{}]{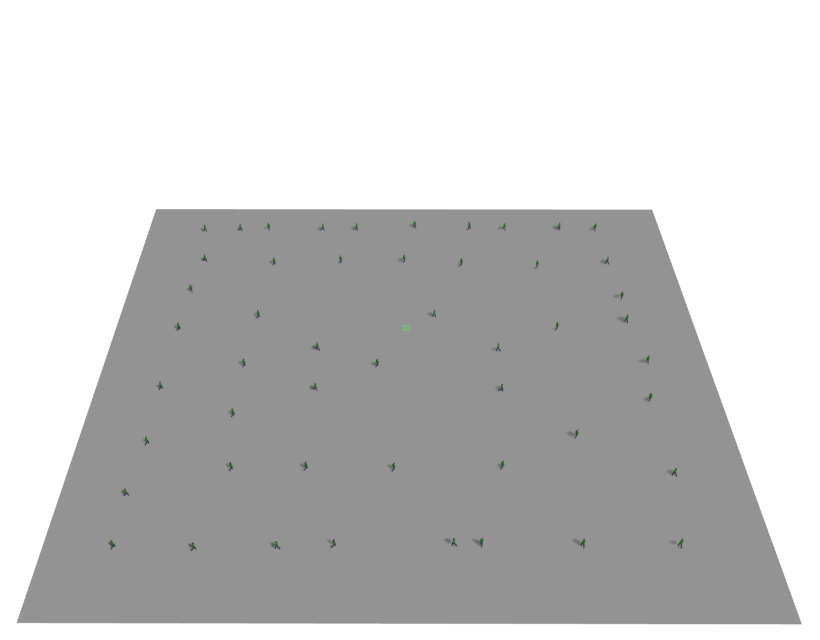}%
	}
	\subfloat{
		\centering
		\includegraphics[width=\simWidth{}]{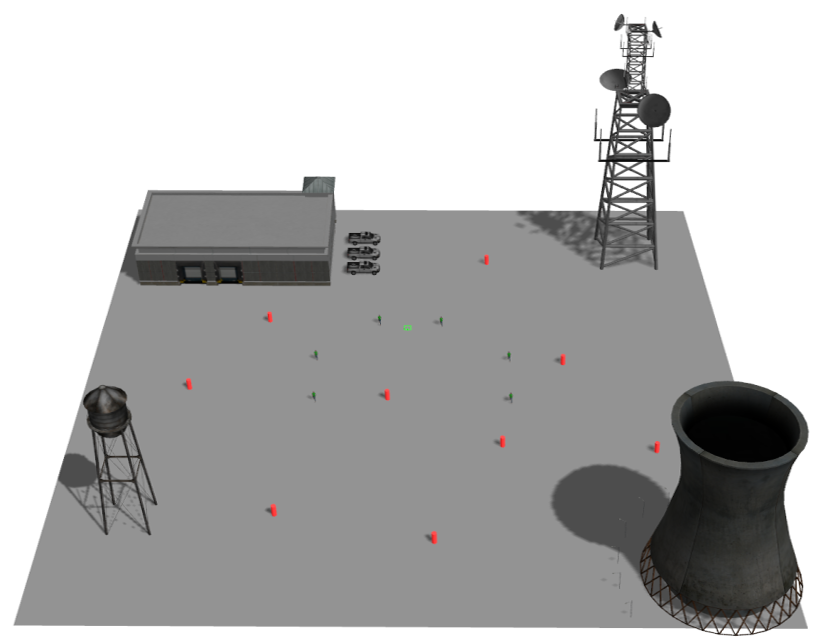}%
	}
	\subfloat{
		\centering
		\includegraphics[width=\simWidth{}]{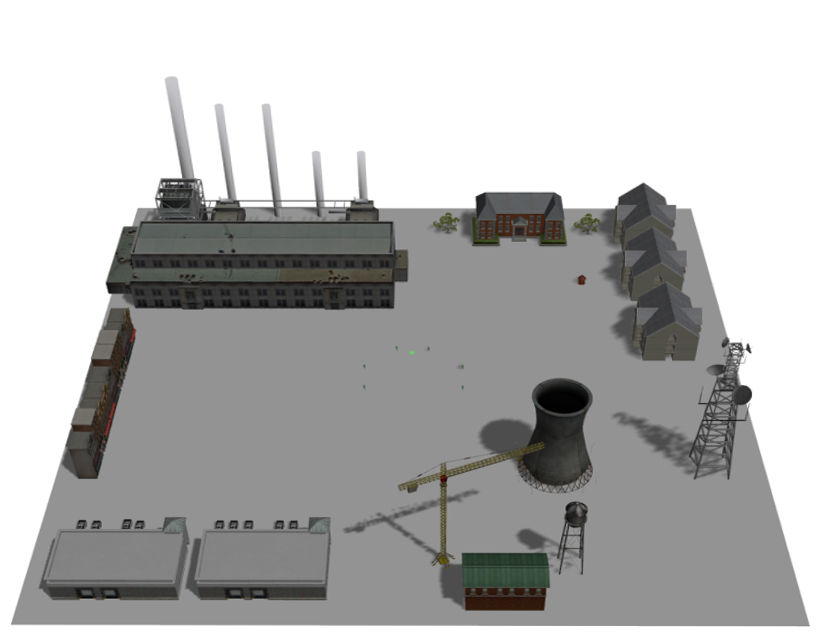}%
	}
	\subfloat{
		\centering
		\includegraphics[width=\simWidth{}]{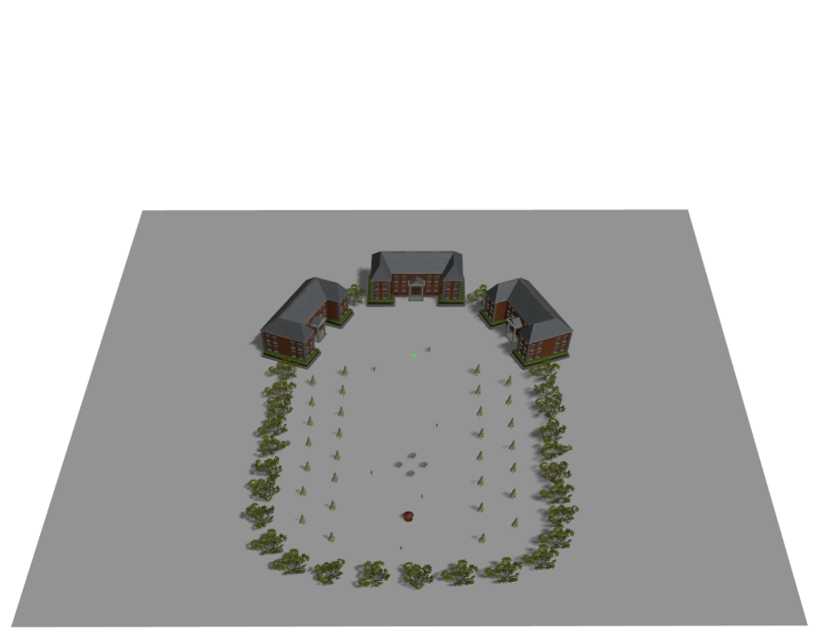}%
	}
	\caption[Simulation Data Examples]{Exemplary views on environments the data sets were recorded in. Left to right: Real-World: Aerial photo~\cite{lbh_maps} of the LBH in Bonn, Germany. Simulated: \textit{Empty field} sequences serve as a baseline providing easier circumstances. \textit{Industrial setting} with six dynamic persons and nine static distractors in red. \textit{Apartments, industry and shops} containing more natural distractors and occluders on a wider field. \textit{Park} with buildings, fountains and a variety of trees providing numerous possibilities for short-term occlusions. }
	\vspace{-0.35cm}
	\label{fig:datasets}
\end{figure*}

\subsubsection{Real World Data}

Our own data set was recorded during flights of a piloted MAV in the courtyard of the \textit{Landesbeh{\"o}rdenhaus} (LBH) in Bonn (\reffig{fig:datasets}).
In the first sequence, only the pilot is visible to the sensor as a dynamic object. 
He moves in a slow pace within an area of about \linebreak \SI{7}{\metre} $\times$ \SI{7}{\metre}. 

The second sequence was recorded with four visible humans. 
They are walking and running, crossing each other's paths, changing speeds and occluding each other. 
The MAV and sensor are flying with velocities of up to $20\,$km/h. 

\subsection{Parameter Optimization}

For parameter optimization, we used \textit{hyperopt}~\cite{bergstra2013hyperopt}, a distributed asynchronous hyperparameter optimization library. 
It utilizes the Tree of Parzen Estimators~\cite{bergstra2011algorithms} to optimize parameters in a specified search space by minimizing a cost function depending on the given parameters---e.g. object size and covariance threshold.

The cost function we utilized is defined by 
\begin{align}
costs = 1 - MOTA.
\end{align}
The costs are equal to zero for a perfect MOTA (\refeq{eq:mot_mota}) of $1.0$ and rising as the MOTA decreases.

\subsection{Quantitative Results}

For quantitative evaluations, we start by comparing the results of our approach to the results reported in~\cite{inlida2017}.
For all evaluations of our method, we only use those hypotheses classified as dynamic at least once. 

The \textit{InLiDa Tracking} approach concentrates on multi person tracking. 
It utilizes global Ensemble of Shape Functions (ESF) descriptors~\cite{wohlkinger2011ensemble} on extracted point clusters and classifies them using random forests into the classes \textit{Person} and \textit{Not person} to generate person detections. 
For tracking, existing hypotheses are matched to their closest detections within a search radius of 0.5m and propagated utilizing a circular velocity buffer. 

The task for evaluation was to track humans only, distinguishing them from the dynamic robot present in four of six sequences.
They evaluated their approach by training parameters on one InLiDa sequence and testing on the remaining. 
Each sequence was used for training once. 

The \textit{InLiDa Tracking} achieves a total MOTA of $-0.213$, our approach evaluated in the same way achieves a total MOTA of $0.071$.
The resulting MOTAs are rather low, considering that using no tracker at all results in a MOTA of zero. 
One possible reason is the utilized evaluation procedure. 
Training on one sequence only increases the risk of overfitting the parameters. 
Additionally, the methods have difficulties to distinguish between the robot and humans, if there is no robot present in the training sequence. 
Applying the methods to other unseen sequences, with an attendant robot, yields bad results.  

We evaluated our method a second time on the InLiDa. 
Due to the limited size of the data set, we perform a Leave-one-out cross-validation (LOOCV).
For this, we split the data set containing $n$ sequences into a training set of size $n-1$ for parameter optimization and a test set consisting of the left out sequence.  
The test set serves the purpose of evaluating the method's performance and ability to generalize on unseen data. 
This process is successively repeated $n$-times, each time leaving another sequence out.
Table \ref{tab:evaluation_results} presents the results of this evaluation on the InLiDa and the simulated data sets.

\begin{table}[t]
	\setlength{\tabcolsep}{5pt}
	\begin{center}
		\caption{Results of quantitative evaluation.}
		\begin{tabular}{l|ccccl}
			Data Set 					& MOTA	& MOTP		& MT	& PT 	& ML \\ 
			\midrule
			InLiDa 						& 0.562 & 0.108m	& 0.49 	& 0.43 	& 0.08 \\ 
			Simulated static sensor 	& 0.677 & 0.044m 	& 0.75 	& 0.25 	& 0.0 \\ 
			Simulated dynamic sensor 	& 0.533 & 0.033m 	& 0.57 	& 0.40 	& 0.03  
		\end{tabular}
		\label{tab:evaluation_results}
	\end{center}
\end{table}

\begin{figure}[t]
	\subfloat{
		\centering
		\resizebox{0.48\linewidth}{!}{%
			\input{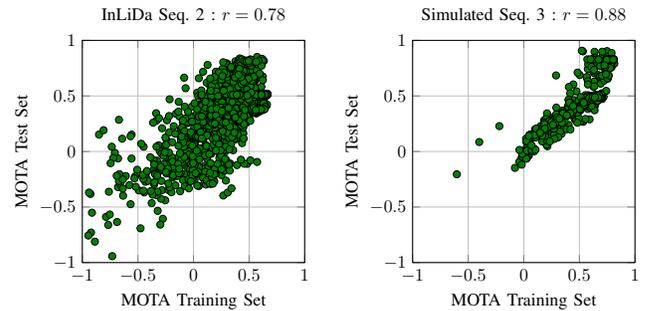}%
		}
	}
	\subfloat{
		\centering
		\resizebox{0.48\linewidth}{!}{%
			\input{Figures/evaluation/correlation/simulation_seq_3.tex}%
		}
	}
	\caption[MOTA Training vs. Test Set on InLiDa]{Plots visualizing the MOTA on the training set against the MOTA on the test sequence during parameter optimization. The titles report the sequence utilized as the test set and the Pearson correlation coefficient $r$.}
	\vspace{-0.35cm}
	\label{fig:correlation_inlida}
\end{figure}

To get a better insight into the method's ability to generalize to unseen data, we plot the MOTAs $\{x_1,\dots,x_n\}$ on the training set against the MOTAs $\{y_1,\dots,y_n\}$ on the test set computed during $n$ runs of the optimization process. 
For each sequence, we additionally compute the Pearson correlation coefficient $r$ defined by 
\begin{align}
r =\frac{\sum_{i=1}^n(x_i - \bar{x})(y_i - \bar{y})}{\sqrt{\strut\sum_{i=1}^n(x_i - \bar{x})^2} \sqrt{\strut\sum_{i=1}^n(y_i - \bar{y})^2}}
\end{align}
with $\bar{x}$ and $\bar{y}$ as the means of the MOTAs on the training and test set, respectively.
The correlation coefficients during the optimizations on the InLiDa sequences range from 0.68 to 0.83, on the simulated sequences from 0.47 to 0.90.
We present the plots corresponding to the test sequences with the median correlation coefficient per data set in \reffig{fig:correlation_inlida}.

\subsection{Qualitative Results}

\begin{figure}[t]
	\subfloat{
		\centering
		\includegraphics[width=0.475\linewidth]{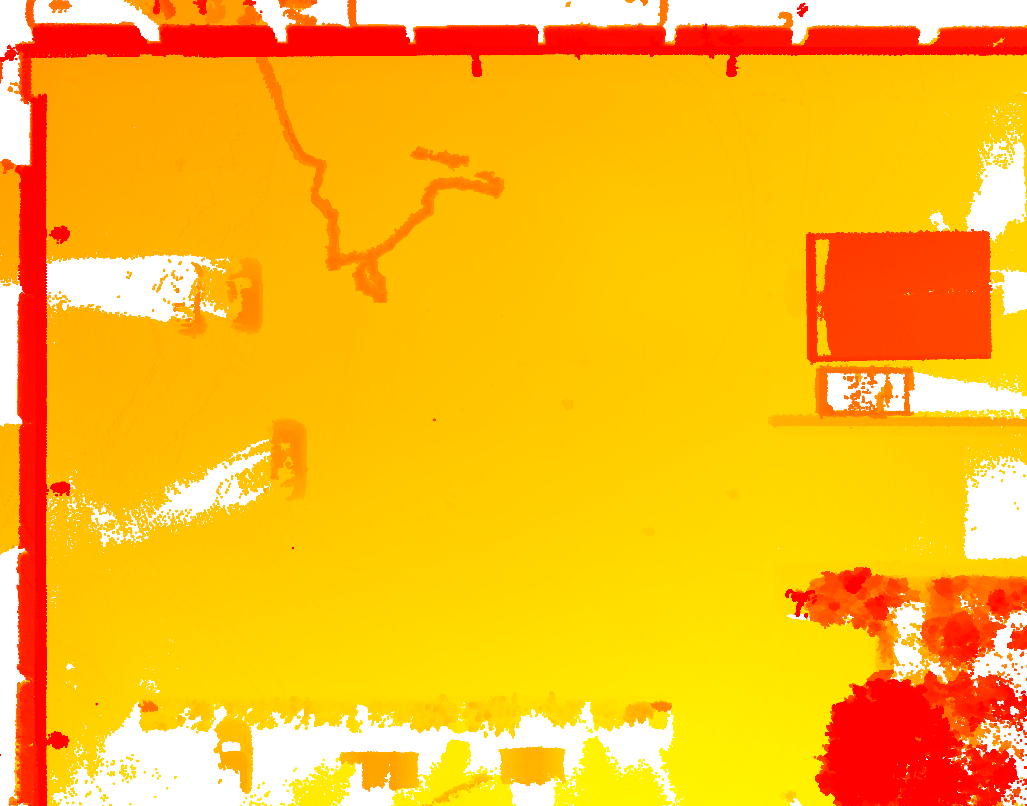}%
	}
	\hfill
	\subfloat{
		\centering
		\includegraphics[width=0.475\linewidth]{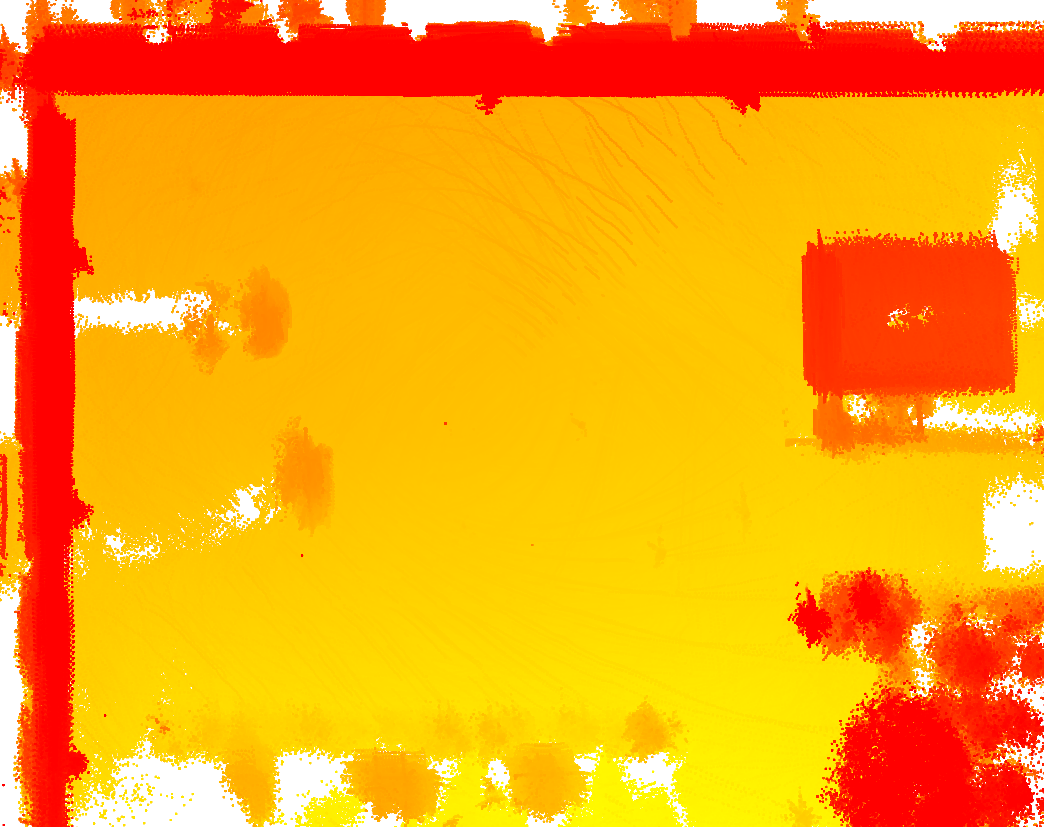}%
	}
	\\[1ex]
	\subfloat{
		\centering
		\includegraphics[width=0.475\linewidth]{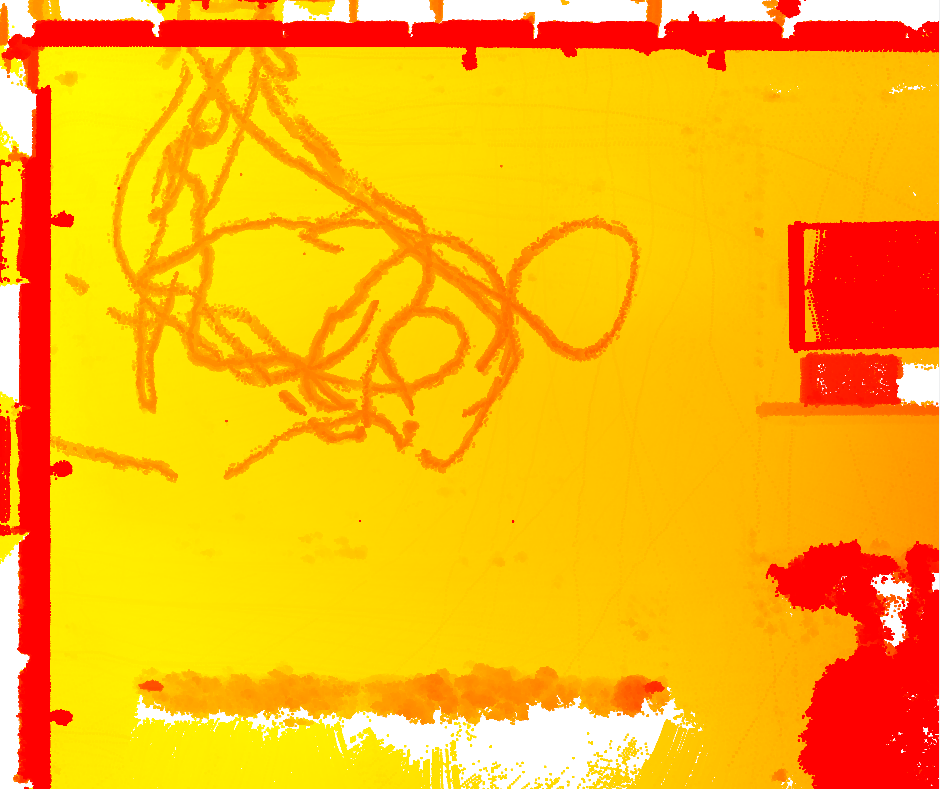}%
	}
	\hfill
	\subfloat{
		\centering
		\includegraphics[width=0.475\linewidth]{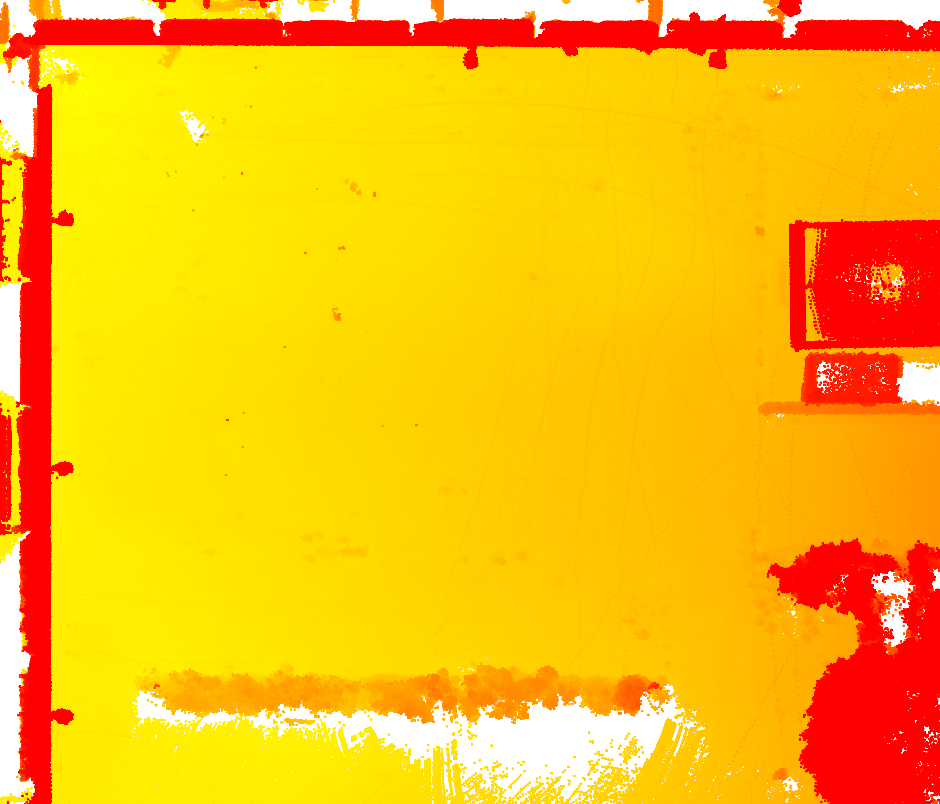}%
	}
	\caption[Qualitative Evaluation]{Top down views of two mapped sequences. Color encodes height (yellow low, red high). Left: Raw measurements mapped into one coordinate frame. Measurements on dynamic objects result in artifacts visible as orange lines on the yellow ground. Right: Same scans with dynamic objects filtered out. Top: Pilot only. Bottom: Four persons.}
	\vspace{-0.35cm}
	\label{fig:quality_maps}
\end{figure}

For a qualitative evaluation, we filter the sequences recorded in the courtyard of the LBH as described in \refsec{sec:object_filter}.
Measurements on dynamic objects create artifacts.
Our method is able to filter out most points corresponding to those objects, even for noisy mapping results present in the first sequence (\reffig{fig:quality_maps}).  

\subsection{Run Time}

\begin{figure}[t]
    \vspace{0.1cm}
	\subfloat{
		\centering
		\input{Figures/evaluation/run_time/run_time_inlida.tex}%
	}
	\vspace{0.1cm}
	\subfloat{
		\centering
		\input{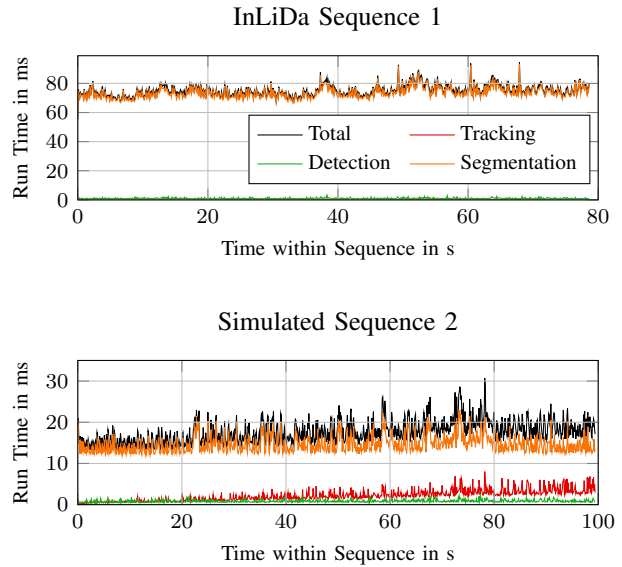}%
	}
	\caption[Run Time Plots]{Run time in milliseconds of method and modules per scan on exemplary sequences. }
	\vspace{-0.35cm}
	\label{fig:run_time}
\end{figure}

Finally, we inspect the real-time capability of our method by plotting the run time per scan for two example sequences (\reffig{fig:run_time}).
We measured the time for each module---Segmentation, Detection and Tracking---to process incoming data. 
We assume a sequential procession of the data on one CPU core. 
In practice, all modules are able to process the data of the next time step directly after processing the current data. 
The method was executed on the hardware of our MAV consisting of an \textit{Intel Core i7-6770HQ} CPU and 32 GB of RAM.

We chose the most demanding sequences from the InLiDa and the simulated data set: a sequence in the hall with a wall close to the sensor resulting in large kernel sizes during segmentation and the simulated sequence with 50 persons present.
Our method processes the data before the next scan is available after 100\,ms.

\subsection{Discussion}

For better results, the application of a more sophisticated but still efficient tracking approach should be investigated. 
Complex movement patterns of humans can only be tracked to a limited extent by the utilized Kalman filter. 

Furthermore, the information about occlusions could help to adjust the time an occluded hypothesis is retained.
The same information can be utilized to more robustly distinguish between static and dynamic objects. 
During partial occlusions, the bounding box of the occluded object changes its shape and size, as the object is only partially visible to the sensor. 
In some instances this is interpreted as a movement of the occluded object. 
Incorporating this information would counteract false classifications and, 
hence, enable the approach to filter dynamic objects more precisely.

Lastly, the run time of the proposed segmentation method depends on the distance to the environment. 
Close objects combined with a large specified target width increase the run time.
Replacing the median filters by an approach that compares the measured distance of a point to the distances of two neighboring background points could generate a similar segmentation while being computationally more efficient.

\section{Conclusion}

We implemented a method for real-time multi object tracking of small objects in the sparse point clouds generated by a Velodyne VLP-16 on the limited hardware of a MAV. 
For this, we proposed a novel segmentation approach to segment point groups of a specified width range in single scan rings and implemented efficient algorithms for detection and tracking utilizing the structure of the data. 
We evaluated our approach on simulated and real in- and outdoor data sets achieving results comparable to the state of the art.
As a practical application, we filter data corresponding to dynamic objects and map the static part of the scene.

\addtolength{\textheight}{-9.5cm}   


\def\bibfont{\footnotesize}
\bibliographystyle{IEEEtranBST/IEEEtran}
\bibliography{bibliography}
\end{document}

%% file: Figures/evaluation/correlation/simulation_seq_3.tex
\begin{tikzpicture}

\begin{axis}[
axis on top,
height=6cm,
tick pos=both,
title={Simulated Seq. 3 : $r = 0.88$},
width=6cm,
xlabel={MOTA Training Set},
xmajorgrids,
xmin=-1, xmax=1,
ylabel={MOTA Test Set},
ymajorgrids,
ymin=-1, ymax=1
]
\addplot [only marks, draw=black, fill=green!50.19607843137255!black, colormap={mymap}{[1pt]
  rgb(0pt)=(0,0,0.5);
  rgb(22pt)=(0,0,1);
  rgb(25pt)=(0,0,1);
  rgb(68pt)=(0,0.86,1);
  rgb(70pt)=(0,0.9,0.967741935483871);
  rgb(75pt)=(0.0806451612903226,1,0.887096774193548);
  rgb(128pt)=(0.935483870967742,1,0.0322580645161291);
  rgb(130pt)=(0.967741935483871,0.962962962962963,0);
  rgb(132pt)=(1,0.925925925925926,0);
  rgb(178pt)=(1,0.0740740740740741,0);
  rgb(182pt)=(0.909090909090909,0,0);
  rgb(200pt)=(0.5,0,0)
}]
table [row sep=\\]{%
x                      y\\ 
+1.285379066601002e-01 +2.039844851904090e-01\\ 
+1.458182450244793e-01 +1.632581100141044e-01\\ 
+1.087082476317390e-01 +1.602609308885754e-01\\ 
-7.940554477244421e-02 -1.477433004231312e-01\\ 
+4.806048842666355e-02 +9.255994358251063e-02\\ 
+4.136127929546047e-01 +3.619534555712270e-01\\ 
-2.074219994785476e-02 -6.082510578279265e-02\\ 
+2.101451374605290e-01 +2.157968970380818e-01\\ 
+2.255279701034218e-02 +8.638928067700991e-02\\ 
+2.281497146499030e-01 +2.431241184767278e-01\\ 
+3.124366291028124e-02 +3.596614950634702e-02\\ 
+2.902169819519684e-01 +2.727433004231312e-01\\ 
+2.257452417509198e-01 +2.875528913963329e-01\\ 
+6.324777658680727e-01 +7.046897038081805e-01\\ 
+2.831773805730176e-02 -5.870944992947824e-02\\ 
+1.139517367247024e-01 +1.450987306064880e-01\\ 
+1.690373417538167e-01 +2.055712270803949e-01\\ 
+4.132506735421071e-02 -3.261636107193233e-02\\ 
+4.525044178568327e-02 -3.808180535966144e-02\\ 
+3.303977519626872e-01 +3.418547249647391e-01\\ 
+3.967959674382224e-01 +3.589562764456982e-01\\ 
+2.122888843825140e-01 +2.254936530324401e-01\\ 
+2.489933080332574e-02 +3.244005641748937e-02\\ 
+2.930994524754483e-01 +1.710155148095910e-01\\ 
+2.564964222602045e-01 +2.413610719322991e-01\\ 
+3.992004403372056e-01 +3.399153737658674e-01\\ 
+5.219009820678466e-01 +2.736248236953456e-01\\ 
+9.438280367333929e-02 +2.380112834978843e-02\\ 
+2.903328601639676e-01 +2.572284908321579e-01\\ 
+2.122743996060141e-01 +4.724964739069115e-02\\ 
+4.716532923896984e-01 +3.912200282087447e-01\\ 
+5.036646484544743e-01 +6.881170662905500e-01\\ 
+2.564674527072047e-01 +2.764456981664316e-01\\ 
+5.916596656913584e-01 +2.893159379407616e-01\\ 
+3.761551609258669e-01 +3.229901269393513e-01\\ 
+5.116023059764188e-01 +4.377644569816643e-01\\ 
+2.320750890813755e-01 +2.614598025387870e-01\\ 
+5.244503027318288e-01 +4.522214386459803e-01\\ 
+3.277180683102060e-01 +1.659026798307476e-01\\ 
+6.322604942205745e-02 +2.133286318758820e-02\\ 
+2.624786349546626e-01 +2.482369534555712e-01\\ 
+3.762565543613662e-01 +3.259873060648801e-01\\ 
+4.180306497870738e-01 +3.515514809590973e-01\\ 
-4.028216344621804e-02 -9.273624823695337e-02\\ 
+1.456444277064805e-01 +1.696050775740480e-01\\ 
+2.022799038210841e-01 +2.011636107193230e-01\\ 
+3.334395550276659e-02 +1.763046544428826e-03\\ 
-6.023928850777833e-01 -2.050423131170662e-01\\ 
+4.465222051623744e-01 +3.541960507757405e-01\\ 
+1.504244039514471e-01 +1.729548660084627e-01\\ 
+3.437237463425935e-02 -4.460507757404786e-02\\ 
+5.730901822184884e-01 +4.984132581100141e-01\\ 
+5.544772444161183e-02 +3.349788434414691e-03\\ 
+2.374199716098381e-01 +1.826516220028209e-01\\ 
+4.420174396709059e-01 +3.485543018335684e-01\\ 
+2.607984008806744e-01 +2.415373765867419e-01\\ 
+1.393435499290241e-02 -5.465444287729193e-02\\ 
+3.392914047336250e-01 +2.730959097320169e-01\\ 
-1.896057243836724e-02 -1.188293370944993e-01\\ 
+6.988904661201079e-02 +2.150916784203105e-02\\ 
-4.008806744111939e-01 +8.515514809590974e-02\\ 
+4.315884005909789e-01 +3.630112834978844e-01\\ 
-1.310872273240826e-02 -4.619181946403383e-02\\ 
+2.278745038964048e-01 +2.180888575458392e-01\\ 
+2.139836032330021e-01 +8.039492242595203e-02\\ 
+5.703091051305078e-01 +4.984132581100141e-01\\ 
+5.798835423969408e-01 +5.063469675599436e-01\\ 
+4.096729337466323e-01 +3.905148095909732e-01\\ 
+7.058431588400591e-01 +8.543723554301834e-01\\ 
+4.744343694776789e-01 +3.799365303244006e-01\\ 
+2.042353486485704e-01 +6.153032440056416e-02\\ 
+5.342564964222603e-01 +2.672778561354020e-01\\ 
+4.504330948173469e-01 +3.705923836389281e-01\\ 
+5.534922796141255e-01 +5.068758815232722e-01\\ 
+3.677105362264260e-01 +3.372708039492243e-01\\ 
+6.193255888061647e-01 +7.267277856135401e-01\\ 
+5.760595614009676e-01 +7.424188998589563e-01\\ 
+4.627741243952606e-01 +3.538434414668548e-01\\ 
+5.604160027810771e-01 +4.335331452750353e-01\\ 
+4.455951794663808e-01 +6.038434414668548e-01\\ 
+1.834496943712158e-01 +2.817348377997179e-01\\ 
+3.286740635591993e-01 +3.231664315937941e-01\\ 
+4.703062081752079e-01 +4.236600846262342e-01\\ 
+1.656334192763406e-01 +9.062059238363895e-02\\ 
+2.952431993974333e-01 +1.562059238363893e-01\\ 
+3.176946029722761e-01 +1.990479548660085e-01\\ 
+1.266259161621136e-01 +1.960507757404796e-01\\ 
+3.270083142617110e-01 +2.711565585331452e-01\\ 
+3.842666357658101e-01 +2.306064880112835e-01\\ 
+1.490918045134564e-01 +1.659026798307476e-01\\ 
+2.100582288015296e-01 +9.837799717912554e-02\\ 
+1.643877284973493e-01 +1.960507757404796e-01\\ 
+2.749645122975752e-01 +2.849083215796897e-01\\ 
+4.569222746893016e-01 +3.626586741889986e-01\\ 
+5.176714273298764e-01 +4.483427362482370e-01\\ 
+8.714041542339002e-02 +1.415726375176305e-01\\ 
+6.564500709754051e-02 +1.318758815232722e-01\\ 
+3.710999739274023e-01 +3.339210155148096e-01\\ 
+4.363104377299458e-01 +3.617771509167842e-01\\ 
+5.682088125380225e-01 +6.602609308885754e-01\\ 
+3.399721892291202e-01 +1.584978843441467e-01\\ 
+2.434021843042962e-01 +2.741537376586742e-01\\ 
+6.590573307453866e-01 +5.005289139633287e-01\\ 
+4.837625655436136e-01 +7.099788434414669e-01\\ 
+2.252382745734238e-02 -6.488011283497874e-02\\ 
+3.280367333932037e-01 +2.946050775740480e-01\\ 
+1.008140444392943e-01 +1.985190409026798e-01\\ 
+3.600336046814797e-01 +2.029266572637518e-01\\ 
+4.539528955068228e-02 +7.898448519040902e-02\\ 
+6.268576725861119e-01 +4.992947813822285e-01\\ 
+7.019612387380862e-01 +4.839562764456982e-01\\ 
+2.776297111735566e-01 +2.552891396332864e-01\\ 
+4.104695964541267e-01 +3.413258110014105e-01\\ 
+5.058228801529592e-01 +3.816995768688294e-01\\ 
+6.416321446160086e-01 +4.841325811001410e-01\\ 
+5.006373301659955e-01 +3.901622002820875e-01\\ 
+3.913641762507605e-01 +3.233427362482370e-01\\ 
+2.713867725021003e-01 +2.440056417489421e-01\\ 
+4.425678611779020e-01 +3.462623413258110e-01\\ 
+2.602769489266781e-01 +2.725669957686883e-01\\ 
+6.791332309742460e-01 +4.964739069111425e-01\\ 
+2.402589878038182e-01 +2.328984485190410e-01\\ 
+4.278513282540051e-01 +3.742947813822285e-01\\ 
+4.524175091978331e-01 +3.453808180535967e-01\\ 
+4.784321677916514e-02 +1.057827926657263e-01\\ 
+7.120860975115154e-01 +4.878349788434415e-01\\ 
+6.865059822126944e-01 +5.007052186177715e-01\\ 
+3.339175526521626e-01 +3.434414668547250e-01\\ 
+9.406413859034157e-02 +1.939351198871675e-03\\ 
+2.828442307135201e-01 +2.838504936530324e-01\\ 
+1.609693212433733e-01 +1.819464033850494e-01\\ 
+4.364697702714447e-01 +3.750000000000000e-01\\ 
+1.571743098003998e-01 +1.701339915373766e-01\\ 
+2.694892667806136e-01 +3.102961918194640e-01\\ 
+2.086532054810394e-01 +3.429125528913963e-01\\ 
+2.854370057070019e-01 +2.882581100141044e-01\\ 
+5.837944320519135e-01 +4.229548660084627e-01\\ 
+2.156058982009907e-01 +2.685119887165022e-01\\ 
+4.057041049856601e-01 +5.807475317348378e-01\\ 
+4.834439004606159e-01 +3.564880112834978e-01\\ 
+4.029664822271792e-01 +2.385401974612130e-01\\ 
+4.764477534111649e-01 +3.774682651622003e-01\\ 
+2.462846548277761e-01 +2.492947813822285e-01\\ 
+3.861062023812972e-01 +3.323342736248237e-01\\ 
+2.887974738549784e-01 +2.764456981664316e-01\\ 
+6.326370984095715e-01 +8.314527503526093e-01\\ 
+6.551464410904140e-01 +4.828984485190410e-01\\ 
+6.354036907210522e-01 +5.031734837799717e-01\\ 
+6.246704713346274e-01 +8.004231311706629e-01\\ 
+6.336510327645645e-01 +4.869534555712270e-01\\ 
+6.042614212462702e-01 +4.100846262341326e-01\\ 
+7.812798748515311e-01 +7.886107193229901e-01\\ 
+4.238390451635331e-01 +3.741184767277856e-01\\ 
+6.886062748051802e-02 +5.236248236953456e-02\\ 
+5.769865870969610e-01 +3.831100141043724e-01\\ 
+7.065818824415540e-01 +4.994710860366713e-01\\ 
+5.313016020162808e-01 +9.010930888575458e-01\\ 
+5.166430081983835e-01 +2.753878702397743e-01\\ 
+1.560589820099075e-01 +1.787729196050776e-01\\ 
+6.230626611431386e-01 +4.707334273624824e-01\\ 
+6.552333497494134e-01 +4.797249647390691e-01\\ 
+5.507836264086445e-01 +4.273624823695346e-01\\ 
+6.904023870911672e-01 +6.269393511988717e-01\\ 
+4.311248877429821e-01 +3.275740479548660e-01\\ 
+1.575798835423969e-01 +1.773624823695346e-01\\ 
+4.604855297082766e-01 +3.566643159379408e-01\\ 
+5.527101016831311e-01 +4.568053596614950e-01\\ 
+4.253020075900229e-01 +3.432651622002821e-01\\ 
+6.441959500564907e-01 +4.888928067700987e-01\\ 
+2.382456038703323e-01 +2.246121297602257e-01\\ 
+6.577247313073959e-01 +4.686177715091678e-01\\ 
+5.640227121295518e-01 +4.139633286318759e-01\\ 
+1.933862510501463e-01 +1.489774330042313e-01\\ 
-9.125409194936651e-04 -1.763046544428715e-03\\ 
+2.467336828992729e-01 +2.690409026798307e-01\\ 
+1.730061705147889e-01 +2.045133991537377e-01\\ 
+6.633882789188563e-01 +6.297602256699577e-01\\ 
+5.116457603059185e-01 +3.758815232722144e-01\\ 
+6.432834091369970e-01 +4.576868829337094e-01\\ 
+4.716388076131985e-01 +3.393864598025388e-01\\ 
+6.857962281641994e-01 +4.987658674188998e-01\\ 
+3.883368579622817e-01 +3.508462623413258e-01\\ 
+4.282858715490020e-01 +3.356840620592384e-01\\ 
+2.688519366146180e-01 +2.401269393511989e-01\\ 
+6.384744633390307e-01 +4.123765867418900e-01\\ 
+4.268663634520119e-01 +3.407968970380818e-01\\ 
+7.200816941394594e-01 +7.967207334273625e-01\\ 
+5.440047510066920e-01 +4.144922425952046e-01\\ 
+3.183609026912715e-01 +2.886107193229901e-01\\ 
+5.071989339204497e-01 +3.771156558533145e-01\\ 
+3.841073032243113e-01 +2.002820874471086e-01\\ 
+5.948173469683362e-01 +3.986248236953456e-01\\ 
+5.534777948376257e-01 +2.926657263751763e-01\\ 
+5.790868796894464e-01 +7.854372355430184e-01\\ 
+4.203337292505577e-01 +4.347672778561354e-01\\ 
+2.052927373330630e-01 +2.043370944992948e-01\\ 
+7.713143486196006e-02 +1.234132581100141e-01\\ 
+3.122917813378134e-02 +1.269393511988715e-02\\ 
+3.133057156928069e-01 +3.155853314527504e-01\\ 
+6.533068744749269e-01 +8.039492242595204e-01\\ 
+4.160752049595875e-01 +3.707686882933710e-01\\ 
+1.504099191749472e-01 +2.267277856135402e-01\\ 
+7.988499087459080e-01 +8.076516220028209e-01\\ 
+3.463454908890756e-01 +3.092383638928068e-01\\ 
+2.766157768185636e-01 +1.854724964739070e-01\\ 
+5.617630869955677e-01 +4.384696755994358e-01\\ 
+5.827080738144210e-01 +4.502820874471086e-01\\ 
+6.778730554187549e-01 +4.887165021156559e-01\\ 
+5.081259596164434e-02 +8.991537376586756e-03\\ 
+4.658593817897390e-01 +3.489069111424542e-01\\ 
+1.763666386627655e-01 +1.819464033850494e-01\\ 
+7.567716330137026e-01 +7.949576868829338e-01\\ 
+1.888814855586778e-01 +2.143864598025388e-01\\ 
+2.579448999101944e-01 +2.697461212976022e-01\\ 
+1.462962426489759e-01 +1.710155148095910e-01\\ 
+6.192097105941655e-01 +4.181946403385050e-01\\ 
+7.872041484399895e-01 +8.044781382228491e-01\\ 
+5.586778296010892e-01 +4.118476727785614e-01\\ 
+5.204525044178568e-01 +4.788434414668548e-01\\ 
+7.046843767200672e-01 +4.728490832157969e-01\\ 
+5.308670587212838e-02 +1.468617771509168e-01\\ 
+2.744430603435789e-01 +2.514104372355430e-01\\ 
+6.691821895188157e-01 +4.851904090267983e-01\\ 
+7.033662620585764e-01 +4.973554301833568e-01\\ 
+4.899185955560705e-01 +4.298307475317349e-01\\ 
+6.245690778991280e-01 +4.647390691114246e-01\\ 
+3.918132043222577e-02 +9.255994358251063e-02\\ 
+2.503838465772473e-01 +2.277856135401974e-01\\ 
+2.795996407775428e-01 +2.909026798307476e-01\\ 
+5.020568382629855e-01 +3.788787023977433e-01\\ 
+6.576812769778962e-01 +4.790197461212976e-01\\ 
+4.184651930820708e-01 +3.422073342736248e-01\\ 
+7.256148787624207e-01 +7.949576868829338e-01\\ 
+5.121961818129146e-01 +3.801128349788434e-01\\ 
+1.254381644891219e-01 +1.599083215796897e-01\\ 
+4.182334366580723e-01 +3.374471086036671e-01\\ 
+6.468466641559720e-01 +4.610366713681241e-01\\ 
+6.110837509777225e-01 +4.382933709449930e-01\\ 
+1.472522378979693e-01 +1.724259520451340e-01\\ 
+3.078449549523451e-01 +2.776798307475318e-01\\ 
+6.424143225470031e-01 +4.689703808180536e-01\\ 
+3.922622323937541e-01 +3.427362482369535e-01\\ 
+2.088125380225383e-01 +2.579337094499294e-01\\ 
+4.391494539239259e-01 +3.437940761636107e-01\\ 
+2.668240679046322e-01 +2.336036671368125e-01\\ 
+6.159940902111880e-01 +4.733779971791255e-01\\ 
+1.114024160607202e-01 +1.711918194640338e-01\\ 
+6.928213447666502e-01 +7.829689703808180e-01\\ 
+6.793794721747444e-01 +6.165373765867419e-01\\ 
+3.978388713462151e-01 +3.462623413258110e-01\\ 
+3.519800689475361e-01 +3.224612129760226e-01\\ 
+5.073148121324488e-01 +3.813469675599436e-01\\ 
+2.531794084417277e-01 +1.110719322990127e-01\\ 
+6.551609258669139e-01 +4.899506346967560e-01\\ 
+2.659984356441381e-01 +2.358956276445698e-01\\ 
+3.598453025869811e-01 +3.374471086036671e-01\\ 
+6.455864886004810e-01 +4.837799717912553e-01\\ 
+7.335525362843651e-01 +7.805007052186178e-01\\ 
+6.361858686520467e-01 +4.753173483779972e-01\\ 
+1.993539789681045e-01 +2.924894217207334e-01\\ 
+6.299863843100901e-01 +5.051128349788434e-01\\ 
+7.361597960543469e-01 +8.298660084626234e-01\\ 
+6.759176105912685e-01 +4.874823695345557e-01\\ 
+6.302471102870884e-01 +4.857193229901270e-01\\ 
+7.483994321967612e-01 +9.033850493653033e-01\\ 
+6.487296851009590e-01 +4.924188998589563e-01\\ 
+6.922709232596541e-01 +6.110719322990127e-01\\ 
+5.113850343289204e-01 +4.365303244005642e-01\\ 
+1.419218401460065e-01 +1.630818053596615e-01\\ 
+7.500217271647498e-01 +8.303949224259520e-01\\ 
+6.643297893913497e-01 +4.957686882933710e-01\\ 
+5.998725339668010e-01 +4.823695345557123e-01\\ 
+4.758683623511689e-01 +4.277150916784203e-01\\ 
+6.570004924824010e-01 +4.841325811001410e-01\\ 
+6.699643674498102e-01 +6.884696755994358e-01\\ 
+5.979750282453142e-01 +4.920662905500706e-01\\ 
+7.318143631043773e-01 +8.187588152327221e-01\\ 
+5.932674758828471e-01 +4.867771509167842e-01\\ 
+4.422202265419044e-01 +3.735895627644570e-01\\ 
+5.639792578000522e-01 +4.372355430183357e-01\\ 
+6.629102812943597e-01 +4.816643159379408e-01\\ 
+7.541643732437209e-01 +8.284555712270804e-01\\ 
+6.602450824183783e-01 +4.917136812411848e-01\\ 
+4.190880384715664e-01 +1.983427362482370e-01\\ 
+7.708798053246039e-01 +8.272214386459802e-01\\ 
+6.435151655609954e-01 +4.693229901269393e-01\\ 
+6.333613372345664e-01 +3.979196050775741e-01\\ 
+4.807787015846345e-01 +3.771156558533145e-01\\ 
+6.692690981778151e-01 +4.888928067700987e-01\\ 
+1.724557490077928e-01 +2.239069111424542e-01\\ 
+6.893160288536748e-01 +4.904795486600846e-01\\ 
+4.536921695298242e-01 +3.457334273624824e-01\\ 
+7.881311741359831e-01 +8.242242595204513e-01\\ 
+2.005417306410963e-01 +2.385401974612130e-01\\ 
+4.732755873576870e-01 +4.298307475317349e-01\\ 
+5.189026333323676e-01 +4.474612129760226e-01\\ 
+4.772009617891596e-01 +3.586036671368125e-01\\ 
+2.762536574060662e-01 +2.441819464033851e-01\\ 
+5.515802891161392e-02 +1.519746121297603e-01\\ 
+6.361134447695472e-01 +8.973906911142454e-01\\ 
+5.517830759871375e-01 +4.231311706629055e-01\\ 
+4.571540311132999e-01 +2.919605077574048e-01\\ 
+2.043801964135694e-02 +1.144217207334274e-01\\ 
+5.669631217590312e-01 +6.710155148095910e-01\\ 
+6.516700947304384e-01 +4.795486600846263e-01\\ 
+6.111127205307222e-01 +4.887165021156559e-01\\ 
+7.435180625162954e-01 +7.864950634696756e-01\\ 
-1.522350010139339e-02 -2.186177715091686e-02\\ 
+5.189460876618675e-01 +4.250705218617772e-01\\ 
+3.258350473652192e-01 +1.780677009873061e-01\\ 
+5.620093281960659e-01 +4.844851904090268e-01\\ 
+4.913381036530606e-01 +3.788787023977433e-01\\ 
+5.645441640835482e-01 +4.326516220028209e-01\\ 
+4.738260088646832e-01 +5.412552891396333e-01\\ 
+4.708711144587039e-01 +3.891043723554302e-01\\ 
+6.314638315130798e-01 +4.719675599435825e-01\\ 
+3.211130102262522e-01 +2.877291960507757e-01\\ 
+1.435875894434949e-01 +1.648448519040903e-01\\ 
+6.397346388945219e-01 +8.464386459802539e-01\\ 
+2.542078275732206e-01 +2.253173483779972e-01\\ 
+4.620064312407659e-01 +3.425599435825106e-01\\ 
+2.723137981980938e-01 +1.528561354019746e-01\\ 
+4.509400619948434e-01 +4.613892806770099e-01\\ 
+2.892609867029750e-02 +8.638928067700946e-03\\ 
+1.832613922767171e-01 +2.616361071932299e-01\\ 
+4.674961615342276e-01 +2.803244005641748e-01\\ 
+3.520235232770358e-01 +3.476727785613540e-01\\ 
+2.508473594252441e-01 +2.327221438645980e-01\\ 
+7.068715779715520e-01 +4.996473906911142e-01\\ 
+7.494568208812538e-01 +8.277503526093088e-01\\ 
+1.808424346012341e-01 +2.029266572637518e-01\\ 
+2.493264578927548e-01 +2.395980253878702e-01\\ 
+4.028940583446797e-01 +4.180183356840621e-01\\ 
+4.762594513166662e-02 +1.205923836389281e-01\\ 
+6.367218053825430e-01 +3.709449929478138e-01\\ 
+7.488484602682581e-01 +6.934062059238364e-01\\ 
+5.609809090645732e-01 +4.240126939351199e-01\\ 
+4.930328225035487e-01 +4.583921015514809e-01\\ 
+6.849850806802051e-01 +6.773624823695346e-01\\ 
+6.188186216286683e-01 +4.881875881523272e-01\\ 
+1.696167328138127e-01 +2.062764456981664e-01\\ 
+5.038964048784726e-01 +3.677715091678421e-01\\ 
+5.590978881195863e-01 +4.342383638928068e-01\\ 
+3.498508068020511e-01 +3.016572637517630e-01\\ 
+7.138097859150039e-02 +1.145980253878698e-02\\ 
+4.173933196210783e-01 +5.232722143864599e-01\\ 
+5.908050638778644e-01 +4.285966149506347e-01\\ 
+5.015933254149889e-01 +3.915726375176305e-01\\ 
+5.564471740201049e-01 +4.411142454160790e-01\\ 
+4.518091485848373e-01 +3.741184767277856e-01\\ 
+3.101190648628291e-01 +1.271156558533145e-01\\ 
+5.591123728960863e-01 +4.684414668547250e-01\\ 
+5.565630522321041e-01 +4.263046544428772e-01\\ 
+5.909499116428634e-01 +6.128349788434415e-01\\ 
+3.173614531127785e-01 +2.656911142454160e-01\\ 
+3.890755815637764e-01 +3.233427362482370e-01\\ 
+6.343607868130594e-01 +4.881875881523272e-01\\ 
+4.245487992120281e-01 +3.429125528913963e-01\\ 
+2.553666096932128e-02 +9.203102961918197e-02\\ 
+7.512819027202411e-01 +7.833215796897038e-01\\ 
+3.981140820997132e-01 +4.069111424541608e-01\\ 
+6.333178829050667e-01 +6.294076163610720e-01\\ 
+4.880935137170833e-01 +3.790550070521862e-01\\ 
+2.764709290535647e-01 +1.135401974612130e-01\\ 
+7.670558243286306e-01 +8.586036671368125e-01\\ 
+5.894869492163736e-01 +4.315937940761636e-01\\ 
+2.337698079318636e-01 +9.062059238363895e-02\\ 
+2.701555664996089e-01 +2.061001410437235e-01\\ 
+5.229004316463397e-01 +4.192524682651622e-01\\ 
+6.181957762391727e-01 +8.298660084626234e-01\\ 
+4.012717633766911e-01 +4.649153737658674e-01\\ 
+4.202033662620586e-01 +4.377644569816643e-01\\ 
+2.852487036125032e-01 +3.046544428772919e-01\\ 
+7.065384281120543e-01 +8.051833568406206e-01\\ 
+4.952345085315334e-01 +3.868124118476728e-01\\ 
+1.907789912801645e-01 +2.032792665726375e-01\\ 
+7.469944088762710e-01 +7.507052186177715e-01\\ 
+3.834120339523162e-01 +3.767630465444288e-01\\ 
+2.797734580955415e-01 +2.452397743300423e-01\\ 
+6.980648338596136e-01 +4.874823695345557e-01\\ 
+5.415857933312089e-01 +4.255994358251057e-01\\ 
+1.051449926127640e-01 +1.569111424541603e-02\\ 
+6.276398505171066e-01 +5.821579689703809e-01\\ 
+6.744401633882789e-01 +4.876586741889986e-01\\ 
+5.118630319534170e-01 +4.164315937940761e-01\\ 
+3.012833511978911e-01 +2.820874471086037e-01\\ 
+4.474492308583679e-01 +3.425599435825106e-01\\ 
+6.778006315362555e-01 +4.899506346967560e-01\\ 
-3.907992699672636e-02 +1.392806770098731e-02\\ 
+6.126625916162114e-01 +4.869534555712270e-01\\ 
+5.221761928213448e-01 +9.065585331452750e-01\\ 
+4.116428633506185e-01 +3.482016925246827e-01\\ 
+5.598221269445812e-01 +4.908321579689704e-01\\ 
+6.337379414235638e-01 +5.862129760225669e-01\\ 
+6.410962078855122e-01 +4.821932299012694e-01\\ 
+1.987601031316086e-01 +2.491184767277856e-01\\ 
+4.513166661838408e-01 +3.887517630465445e-01\\ 
+6.043193603522697e-01 +4.448166431593794e-01\\ 
+5.915872418088589e-01 +4.541607898448519e-01\\ 
+5.413685216837104e-01 +4.844851904090268e-01\\ 
+5.764651351429648e-01 +6.329337094499294e-01\\ 
+8.142037718358006e-01 +7.903737658674189e-01\\ 
+7.064370346765549e-01 +4.978843441466855e-01\\ 
+7.977056114024160e-01 +8.101198871650211e-01\\ 
+6.685303745763203e-01 +4.897743300423131e-01\\ 
+7.956487731394305e-01 +8.111777150916784e-01\\ 
+6.906920826211651e-01 +5.021156558533145e-01\\ 
+7.727483414930908e-01 +7.857898448519041e-01\\ 
+6.961673281381269e-01 +4.989421720733427e-01\\ 
+7.890147455024769e-01 +7.572284908321579e-01\\ 
+6.650974825458443e-01 +4.962976022566996e-01\\ 
+6.829137576407196e-01 +4.977080394922426e-01\\ 
+7.889568063964774e-01 +8.869887165021156e-01\\ 
+7.242533097714302e-01 +6.382228490832158e-01\\ 
+6.488310785364582e-01 +4.938293370944993e-01\\ 
+7.840319823865118e-01 +8.111777150916784e-01\\ 
+6.994843419566036e-01 +5.022919605077574e-01\\ 
+7.351313769228540e-01 +7.224964739069111e-01\\ 
+7.070309105130508e-01 +5.126939351198871e-01\\ 
+6.905182653031663e-01 +5.040550070521862e-01\\ 
+7.055244937570613e-01 +8.977433004231312e-01\\ 
+6.976447753411165e-01 +5.054654442877292e-01\\ 
+7.865813030504939e-01 +7.997179125528914e-01\\ 
+5.684695385150207e-01 +4.388222849083215e-01\\ 
+6.838552681132131e-01 +4.802538787023978e-01\\ 
+6.677047423158260e-01 +4.964739069111425e-01\\ 
+6.684579506938209e-01 +5.033497884344147e-01\\ 
+6.084330368782409e-01 +7.616361071932299e-01\\ 
+7.108404067325241e-01 +5.049365303244006e-01\\ 
+5.733943625249862e-01 +4.375881523272215e-01\\ 
+7.038297749065732e-01 +4.934767277856136e-01\\ 
+6.588690286508879e-01 +4.929478138222849e-01\\ 
+6.692546134013152e-01 +5.003526093088857e-01\\ 
+7.709087748776037e-01 +7.912552891396333e-01\\ 
+6.748747066832759e-01 +4.929478138222849e-01\\ 
+5.819403806599264e-01 +4.356488011283498e-01\\ 
+6.026246415017816e-01 +4.532792665726375e-01\\ 
+7.014687563370897e-01 +5.236248236953456e-01\\ 
+6.571163706944002e-01 +4.906558533145275e-01\\ 
+7.801790318375388e-01 +7.547602256699577e-01\\ 
+1.953706654306324e-01 +2.133286318758815e-01\\ 
+5.038094962194734e-01 +3.922778561354020e-01\\ 
+2.647961991946465e-01 +2.455923836389281e-01\\ 
+5.825921956024218e-01 +4.397038081805360e-01\\ 
+5.023465337929836e-01 +3.661847672778561e-01\\ 
+1.915901387641589e-01 +2.022214386459803e-01\\ 
+7.555549117877112e-01 +6.565585331452750e-01\\ 
+6.384165242330311e-01 +4.876586741889986e-01\\ 
+7.004113676525971e-01 +5.022919605077574e-01\\ 
+6.305223210405864e-01 +7.355430183356841e-01\\ 
+2.630435412381587e-01 +2.364245416078985e-01\\ 
+5.603870332280774e-01 +4.407616361071932e-01\\ 
+4.945537240360381e-01 +3.004231311706629e-01\\ 
+8.059764187838582e-01 +8.325105782792666e-01\\ 
+7.047857701555664e-01 +4.888928067700987e-01\\ 
+6.492945913844550e-01 +4.888928067700987e-01\\ 
+7.114487673455199e-01 +7.960155148095910e-01\\ 
+7.001796112285987e-01 +5.040550070521862e-01\\ 
+4.371650395434399e-01 +3.314527503526093e-01\\ 
+2.539760711492222e-01 +2.344851904090268e-01\\ 
+6.668935948318317e-01 +5.007052186177715e-01\\ 
+2.063646107940554e-01 +2.424188998589563e-01\\ 
+3.158115820272893e-01 +2.928420310296191e-01\\ 
-2.195892117384628e-01 +2.281382228490832e-01\\ 
+5.607346678640748e-01 +4.405853314527504e-01\\ 
+6.768301515107622e-01 +4.925952045133991e-01\\ 
+3.474897882325676e-01 +3.670662905500706e-01\\ 
+1.746139807062776e-01 +1.882933709449930e-01\\ 
+3.820504649613257e-01 +3.210507757404796e-01\\ 
+7.235580404994351e-01 +8.013046544428772e-01\\ 
+6.687186766708190e-01 +4.998236953455572e-01\\ 
+2.921434572264550e-01 +3.349788434414669e-01\\ 
+6.506416755989455e-01 +4.970028208744711e-01\\ 
+5.302297285552884e-01 +4.218970380818053e-01\\ 
+7.746023928850778e-01 +8.065937940761636e-01\\ 
+3.677974448854254e-01 +2.718617771509168e-01\\ 
+5.079521422984443e-01 +4.076163610719323e-01\\ 
+4.923520380080535e-01 +3.758815232722144e-01\\ 
+6.641559720733510e-01 +4.820169252468265e-01\\ 
+4.441756713693907e-01 +3.478490832157969e-01\\ 
+2.903183753874677e-01 +6.844146685472496e-01\\ 
+6.163272400706857e-01 +4.525740479548660e-01\\ 
+6.230047220371390e-01 +5.017630465444287e-01\\ 
+8.082215591413424e-01 +8.309238363892807e-01\\ 
+6.541904458414207e-01 +4.982369534555712e-01\\ 
+4.938294852110432e-01 +3.797602256699577e-01\\ 
+5.444103247486891e-01 +4.330042313117066e-01\\ 
+4.809814884556332e-01 +3.940409026798307e-01\\ 
+6.088530953967380e-01 +4.095557122708039e-01\\ 
+7.414467394768098e-01 +5.456629055007052e-01\\ 
+4.659462904487384e-01 +3.510225669957687e-01\\ 
+7.565109070367044e-01 +8.298660084626234e-01\\ 
+2.412004982763116e-01 +2.339562764456982e-01\\ 
+6.716880558532982e-01 +4.943582510578279e-01\\ 
+5.474955821431675e-01 +4.435825105782792e-01\\ 
+6.784669312552507e-01 +4.904795486600846e-01\\ 
+7.870158463454908e-01 +8.284555712270804e-01\\ 
+7.178365537819751e-01 +4.929478138222849e-01\\ 
+5.716272197919986e-01 +8.295133991537377e-01\\ 
+3.189692633042672e-01 +2.889633286318759e-01\\ 
+4.367304962484428e-01 +3.834626234132581e-01\\ 
+7.026275384570816e-01 +4.973554301833568e-01\\ 
+1.056519597902604e-01 +1.260578279266573e-01\\ 
+4.087893623801385e-01 +4.409379407616361e-01\\ 
+3.735189316028854e-01 +3.372708039492243e-01\\ 
+6.707030910513051e-01 +4.962976022566996e-01\\ 
+4.796488890176425e-01 +3.487306064880112e-01\\ 
+6.753237347547727e-01 +4.917136812411848e-01\\ 
+5.035632550189750e-01 +3.809943582510579e-01\\ 
+6.334627306700658e-01 +4.985895627644570e-01\\ 
+8.041513369448710e-01 +8.302186177715092e-01\\ 
+2.717054375850980e-01 +2.443582510578279e-01\\ 
+9.783018048031522e-02 +1.472143864598026e-01\\ 
+5.560995393841073e-01 +4.388222849083215e-01\\ 
+6.652857846403430e-01 +5.003526093088857e-01\\ 
+1.113010226252209e-01 +2.062764456981664e-01\\ 
+6.874909470146875e-01 +7.679830747531735e-01\\ 
+7.941133868304412e-01 +8.256346967559943e-01\\ 
+2.297575248413917e-01 +2.745063469675599e-01\\ 
+6.456733972594804e-01 +4.862482369534555e-01\\ 
+5.091978330774356e-01 +3.820521861777151e-01\\ 
+4.120629218691155e-01 +3.422073342736248e-01\\ 
+1.462238187664764e-01 +1.710155148095910e-01\\ 
+6.581882441553927e-01 +4.867771509167842e-01\\ 
+1.902430545496683e-01 +1.375176304654443e-01\\ 
+6.739911353167820e-01 +4.901269393511989e-01\\ 
+2.881456589124830e-01 +2.743300423131171e-01\\ 
+7.029896578695791e-01 +6.824753173483780e-01\\ 
+6.810017671427330e-01 +4.906558533145275e-01\\ 
+5.599524899330803e-01 +4.411142454160790e-01\\ 
+6.643008198383499e-01 +4.569816643159379e-01\\ 
+7.521220197572351e-01 +6.581452750352610e-01\\ 
+6.690663113068165e-01 +4.908321579689704e-01\\ 
+3.729974796488890e-01 +3.217559943582511e-01\\ 
+6.887945768996784e-01 +4.959449929478138e-01\\ 
+6.445146151394884e-01 +8.277503526093088e-01\\ 
+5.081838987224427e-01 +3.787023977433004e-01\\ 
+5.988296300588082e-01 +4.740832157968971e-01\\ 
+5.673542107245285e-01 +4.412905500705219e-01\\ 
+4.397578145369216e-01 +3.513751763046544e-01\\ 
+3.426663576581014e-01 +2.566995768688294e-01\\ 
+5.077783249804455e-01 +4.074400564174894e-01\\ 
+1.663576581013355e-01 +1.426304654442877e-01\\ 
+4.492887974738550e-01 +3.751763046544428e-01\\ 
+2.519771719922361e-01 +2.344851904090268e-01\\ 
+6.008574987687940e-01 +8.302186177715092e-01\\ 
+5.189316028853674e-01 +4.139633286318759e-01\\ 
+4.561256119818069e-02 +1.408674188998590e-01\\ 
+4.808076711376343e-01 +2.595204513399154e-01\\ 
+8.001825081838987e-01 +8.312764456981665e-01\\ 
+6.825950925577218e-01 +4.941819464033851e-01\\ 
+5.763782264839654e-01 +6.031382228490831e-01\\ 
+6.631710072713578e-01 +4.911847672778561e-01\\ 
+6.971233233871201e-01 +4.945345557122708e-01\\ 
+7.291201946753962e-01 +7.884344146685472e-01\\ 
+2.474434369477679e-01 +2.648095909732017e-01\\ 
+2.840609519395115e-01 +2.563469675599436e-01\\ 
+2.924910918624526e-01 +3.952750352609309e-01\\ 
+7.663315855036357e-01 +8.296897038081805e-01\\ 
+2.199368463744604e-01 +2.223201692524682e-01\\ 
+5.050986413279643e-01 +3.816995768688294e-01\\ 
+4.192184014600655e-01 +3.434414668547250e-01\\ 
+6.535965700049249e-01 +4.881875881523272e-01\\ 
+8.094093108143341e-01 +8.286318758815232e-01\\ 
+1.584200005793911e-01 +1.875881523272215e-01\\ 
+3.518786755120369e-01 +3.503173483779972e-01\\ 
+1.850140502332049e-01 +2.464739069111425e-01\\ 
+3.520235232770358e-01 +3.436177715091678e-01\\ 
+6.459341232364785e-01 +5.095204513399154e-01\\ 
+4.443784582403894e-01 +4.093794076163610e-01\\ 
+8.069613835858512e-01 +8.265162200282088e-01\\ 
+3.027028592948811e-01 +3.069464033850494e-01\\ 
+6.120976853327154e-01 +4.869534555712270e-01\\ 
+3.869463194182914e-01 +3.251057827926658e-01\\ 
+3.211419797792520e-01 +2.953102961918195e-01\\ 
+6.725861119962919e-01 +8.273977433004231e-01\\ 
+6.266259161621136e-01 +4.846614950634697e-01\\ 
+8.101770039688287e-01 +8.249294781382228e-01\\ 
+7.959964077754280e-01 +8.265162200282088e-01\\ 
+7.593644080071844e-01 +8.238716502115656e-01\\ 
+6.971522929401199e-01 +5.109308885754584e-01\\ 
+7.897534691039717e-01 +8.259873060648801e-01\\ 
+6.849995654567049e-01 +4.880112834978844e-01\\ 
+7.947941713259364e-01 +8.244005641748942e-01\\ 
+6.731799878327878e-01 +5.003526093088857e-01\\ 
+7.891306237144761e-01 +8.102961918194640e-01\\ 
+6.568411599409021e-01 +4.733779971791255e-01\\ 
+7.768185636895623e-01 +8.231664315937941e-01\\ 
+7.647382600886469e-01 +8.254583921015515e-01\\ 
+6.701671543208088e-01 +5.035260930888575e-01\\ 
+6.927634056606506e-01 +4.814880112834978e-01\\ 
+7.800052145195400e-01 +8.236953455571228e-01\\ 
+5.724528520524929e-01 +4.525740479548660e-01\\ 
+6.521770619079348e-01 +4.682651622002821e-01\\ 
+7.628697239201599e-01 +8.254583921015515e-01\\ 
+6.642863350618500e-01 +5.074047954866008e-01\\ 
+7.763695356180654e-01 +7.799717912552891e-01\\ 
+5.882267736608824e-01 +4.814880112834978e-01\\ 
+6.778440858657551e-01 +4.837799717912553e-01\\ 
+7.285552883919001e-01 +7.498236953455570e-01\\ 
+6.143862800196993e-01 +4.673836389280677e-01\\ 
+6.821895188157246e-01 +4.818406205923836e-01\\ 
+7.647527448651468e-01 +6.844146685472496e-01\\ 
+6.944870940641386e-01 +4.860719322990127e-01\\ 
+7.382311190938324e-01 +7.775035260930889e-01\\ 
+6.370259856890408e-01 +4.853667136812412e-01\\ 
+7.722848286450941e-01 +7.822637517630465e-01\\ 
+6.813349170022307e-01 +4.855430183356840e-01\\ 
+6.359106578985486e-01 +4.786671368124118e-01\\ 
+5.469017063066717e-01 +4.303596614950634e-01\\ 
+7.815695703815291e-01 +7.884344146685472e-01\\ 
+5.120947883774154e-01 +4.895980253878702e-01\\ 
+7.329152061183696e-01 +7.708039492242595e-01\\ 
+6.772357252527593e-01 +4.820169252468265e-01\\ 
+6.323039485500739e-01 +4.769040902679831e-01\\ 
+7.746748167675772e-01 +7.515867418899859e-01\\ 
+6.343607868130594e-01 +4.786671368124118e-01\\ 
+1.887945768996785e-01 +2.052186177715092e-01\\ 
+6.789304441032475e-01 +4.853667136812412e-01\\ 
+5.283611923868015e-01 +4.368829337094500e-01\\ 
+7.392305686723254e-01 +6.958744710860367e-01\\ 
+6.391987021640256e-01 +4.700282087447109e-01\\ 
+7.972710681074191e-01 +8.270451339915373e-01\\ 
+5.311857238042816e-01 +4.509873060648801e-01\\ 
+6.651119673223442e-01 +4.573342736248237e-01\\ 
+7.760798400880674e-01 +8.254583921015515e-01\\ 
+6.797705611402416e-02 +5.694640338504942e-02\\ 
+6.265679770561140e-01 +4.823695345557123e-01\\ 
+7.481966453257627e-01 +8.254583921015515e-01\\ 
+6.831165445117182e-01 +4.851904090267983e-01\\ 
+6.417335380515079e-01 +4.836036671368125e-01\\ 
+1.470929053564703e-01 +1.673131170662906e-01\\ 
+5.414119760132101e-01 +4.206629055007052e-01\\ 
+7.326399953648715e-01 +7.618124118476728e-01\\ 
+6.769605144992612e-01 +5.052891396332864e-01\\ 
+2.632897824386570e-01 +2.404795486600846e-01\\ 
+7.255569396564211e-01 +7.757404795486601e-01\\ 
+6.315797097250789e-01 +4.777856135401974e-01\\ 
+8.048466062168661e-01 +8.270451339915373e-01\\ 
+6.215127900576494e-01 +4.894217207334274e-01\\ 
+6.685883136823199e-01 +4.862482369534555e-01\\ 
+5.394275616327240e-01 +4.389985895627645e-01\\ 
+6.398650018830210e-01 +4.866008462623413e-01\\ 
};
\end{axis}

\end{tikzpicture}

%% file: Figures/evaluation/run_time/run_time_inlida.tex
\begin{tikzpicture}

\definecolor{color1}{rgb}{0.976470588235294,0.450980392156863,0.0235294117647059}
\definecolor{color0}{rgb}{0.0823529411764706,0.690196078431373,0.101960784313725}

\begin{axis}[
axis on top,
height=3.5cm,
legend cell align={left},
legend entries={{Total},{Tracking},{Detection},{Segmentation}},
legend columns=2,
legend style={
	at={(0.985,0.35)}, 
	anchor=east, 
	/tikz/column 2/.style={
		column sep=5pt,
	}	
},
tick pos=both,
title={InLiDa Sequence 1},
title style={font=\normalsize},
font=\footnotesize,
width=8.5cm,
xlabel={Time within Sequence in s},
xmajorgrids,
xmin=0, xmax=80,
ylabel={Run Time in ms},
ymajorgrids,
ymin=0, ymax=99
]
\addlegendimage{no markers, black}
\addlegendimage{no markers, red!89.80392156862746!black}
\addlegendimage{no markers, color0}
\addlegendimage{no markers, color1}
\addplot [black]
table [row sep=\\]{%
0	55.927 \\
0.1	76.142 \\
0.2	72.985 \\
0.3	70.325 \\
0.4	73.685 \\
0.5	72.535 \\
0.6	72.634 \\
0.7	72.785 \\
0.8	72.887 \\
0.9	68.468 \\
1	76.185 \\
1.1	71.698 \\
1.2	70.579 \\
1.3	73.182 \\
1.4	69.555 \\
1.5	71.363 \\
1.6	74.422 \\
1.7	70.766 \\
1.8	70.426 \\
1.9	78.715 \\
2	72.51 \\
2.1	76.458 \\
2.2	77.41 \\
2.3	81.658 \\
2.4	74.958 \\
2.5	71.654 \\
2.6	74.588 \\
2.7	72.063 \\
2.8	74.38 \\
2.9	71.183 \\
3	73.837 \\
3.1	73.446 \\
3.2	71.203 \\
3.3	71.188 \\
3.4	72.591 \\
3.5	73.954 \\
3.6	69.44 \\
3.7	77.091 \\
3.8	77.681 \\
3.9	71.688 \\
4	71.365 \\
4.1	72.132 \\
4.2	75.551 \\
4.3	72.061 \\
4.4	69.152 \\
4.5	71.09 \\
4.6	71.913 \\
4.7	71.648 \\
4.8	71.9 \\
4.9	68.932 \\
5	71.771 \\
5.1	70.024 \\
5.2	73.052 \\
5.3	68.946 \\
5.4	69.364 \\
5.5	70.88 \\
5.6	70.467 \\
5.7	70.548 \\
5.8	70.477 \\
5.9	68.834 \\
6	68.917 \\
6.1	69.582 \\
6.2	72.101 \\
6.3	72.135 \\
6.4	70.34 \\
6.5	67.92 \\
6.6	68.08 \\
6.7	68.901 \\
6.8	70.758 \\
6.9	69.853 \\
7	76.079 \\
7.1	70.151 \\
7.2	69.013 \\
7.3	71.908 \\
7.4	70.942 \\
7.5	69.458 \\
7.6	69.167 \\
7.7	70.536 \\
7.8	69.417 \\
7.9	71.531 \\
8	70.014 \\
8.1	69.016 \\
8.2	69.515 \\
8.3	68.315 \\
8.4	70.146 \\
8.5	71.221 \\
8.6	68.971 \\
8.7	68.515 \\
8.8	70.561 \\
8.9	71.865 \\
9	73.777 \\
9.1	76.473 \\
9.2	73.204 \\
9.3	71.769 \\
9.4	74.817 \\
9.5	73.024 \\
9.6	74.451 \\
9.7	71.556 \\
9.8	71.632 \\
9.9	75.832 \\
10	71.961 \\
10.1	70.563 \\
10.2	70.67 \\
10.3	72.008 \\
10.4	72.14 \\
10.5	75.092 \\
10.6	70.711 \\
10.7	71.686 \\
10.8	74.785 \\
10.9	72.204 \\
11	77.855 \\
11.1	70.801 \\
11.2	73.556 \\
11.3	74.534 \\
11.4	72.04 \\
11.5	72.835 \\
11.6	75.814 \\
11.7	73.083 \\
11.8	72.1 \\
11.9	74.553 \\
12	71.602 \\
12.1	75.441 \\
12.2	72.591 \\
12.3	75.679 \\
12.4	76.153 \\
12.5	74.853 \\
12.6	79.084 \\
12.7	76.851 \\
12.8	80.224 \\
12.9	78.697 \\
13	78.056 \\
13.1	78.961 \\
13.2	78.452 \\
13.3	78.934 \\
13.4	76.365 \\
13.5	80.044 \\
13.6	76.97 \\
13.7	78.559 \\
13.8	76.341 \\
13.9	75.823 \\
14	74.222 \\
14.1	73.247 \\
14.2	73.738 \\
14.3	74.048 \\
14.4	72.475 \\
14.5	78.861 \\
14.6	75.098 \\
14.7	74.5 \\
14.8	78.004 \\
14.9	76.825 \\
15	73.36 \\
15.1	73.941 \\
15.2	70.66 \\
15.3	74.2 \\
15.4	73.667 \\
15.5	73.513 \\
15.6	71.198 \\
15.7	74.448 \\
15.8	75.713 \\
15.9	74.456 \\
16	73.947 \\
16.1	71.369 \\
16.2	74.39 \\
16.3	72.473 \\
16.4	74.82 \\
16.5	71.367 \\
16.6	71.452 \\
16.7	77.907 \\
16.8	74.355 \\
16.9	74.268 \\
17	76.093 \\
17.1	74.652 \\
17.2	79.055 \\
17.3	75.886 \\
17.4	78.913 \\
17.5	74.247 \\
17.6	78.095 \\
17.7	76.871 \\
17.8	78.288 \\
17.9	74.365 \\
18	76.226 \\
18.1	77.774 \\
18.2	79.277 \\
18.3	75.39 \\
18.4	77.132 \\
18.5	73.901 \\
18.6	77.935 \\
18.7	74.989 \\
18.8	72.414 \\
18.9	79.152 \\
19	77.004 \\
19.1	72.401 \\
19.2	75.027 \\
19.3	75.988 \\
19.4	71.802 \\
19.5	74.916 \\
19.6	75.376 \\
19.7	69.898 \\
19.8	78.735 \\
19.9	77.574 \\
20	76.881 \\
20.1	73.897 \\
20.2	76.941 \\
20.3	74.046 \\
20.4	75.655 \\
20.5	71.293 \\
20.6	71.296 \\
20.7	70.728 \\
20.8	75.56 \\
20.9	73.297 \\
21	71.538 \\
21.1	73.092 \\
21.2	73.255 \\
21.3	76.206 \\
21.4	71.562 \\
21.5	71.95 \\
21.6	74.214 \\
21.7	74.703 \\
21.8	71.491 \\
21.9	69.985 \\
22	70.601 \\
22.1	74.424 \\
22.2	71.394 \\
22.3	69.526 \\
22.4	70.385 \\
22.5	76.072 \\
22.6	79.137 \\
22.7	72.274 \\
22.8	71.065 \\
22.9	72.451 \\
23	78.35 \\
23.1	69.331 \\
23.2	69.469 \\
23.3	74.049 \\
23.4	70.936 \\
23.5	70.9 \\
23.6	71.693 \\
23.7	71.237 \\
23.8	73.028 \\
23.9	72.961 \\
24	71.239 \\
24.1	72.583 \\
24.2	74.916 \\
24.3	71.28 \\
24.4	69.596 \\
24.5	73.701 \\
24.6	71.474 \\
24.7	76.589 \\
24.8	69.763 \\
24.9	70.359 \\
25	72.247 \\
25.1	76.015 \\
25.2	72.225 \\
25.3	73.689 \\
25.4	69.844 \\
25.5	73.649 \\
25.6	73.485 \\
25.7	70.961 \\
25.8	71.526 \\
25.9	77.283 \\
26	72.974 \\
26.1	70.856 \\
26.2	77.435 \\
26.3	73.412 \\
26.4	72.848 \\
26.5	76.3 \\
26.6	75.047 \\
26.7	73.643 \\
26.8	73.722 \\
26.9	73.279 \\
27	69.78 \\
27.1	74.154 \\
27.2	75.506 \\
27.3	71.081 \\
27.4	71.571 \\
27.5	76.054 \\
27.6	73.116 \\
27.7	72.787 \\
27.8	73.711 \\
27.9	71.789 \\
28	70.325 \\
28.1	72.493 \\
28.2	71.091 \\
28.3	71.211 \\
28.4	73.82 \\
28.5	71.874 \\
28.6	73.214 \\
28.7	77.579 \\
28.8	75.727 \\
28.9	71.85 \\
29	70.141 \\
29.1	71.826 \\
29.2	76.808 \\
29.3	73.512 \\
29.4	74.946 \\
29.5	74.212 \\
29.6	71.13 \\
29.7	81.299 \\
29.8	70.895 \\
29.9	71.019 \\
30	76.898 \\
30.1	72.918 \\
30.2	71.982 \\
30.3	74.292 \\
30.4	73.122 \\
30.5	67.987 \\
30.6	71.41 \\
30.7	72.865 \\
30.8	70.587 \\
30.9	77.695 \\
31	74.202 \\
31.1	73.978 \\
31.2	73.592 \\
31.3	72.321 \\
31.4	74.789 \\
31.5	72.389 \\
31.6	72.777 \\
31.7	73.627 \\
31.8	71.392 \\
31.9	71.264 \\
32	73.776 \\
32.1	70.408 \\
32.2	69.031 \\
32.3	68.85 \\
32.4	73.177 \\
32.5	70.411 \\
32.6	72.172 \\
32.7	70.804 \\
32.8	69.121 \\
32.9	69.928 \\
33	71.842 \\
33.1	67.81 \\
33.2	69.449 \\
33.3	68.273 \\
33.4	72.876 \\
33.5	73.518 \\
33.6	70.774 \\
33.7	70.418 \\
33.8	68.516 \\
33.9	69.414 \\
34	71.079 \\
34.1	70.324 \\
34.2	71.368 \\
34.3	68.904 \\
34.4	70.363 \\
34.5	71.351 \\
34.6	73.319 \\
34.7	77.021 \\
34.8	70.302 \\
34.9	74.504 \\
35	73.824 \\
35.1	74.403 \\
35.2	76.062 \\
35.3	69.912 \\
35.4	71.226 \\
35.5	72.932 \\
35.6	70.6 \\
35.7	71.422 \\
35.8	70.054 \\
35.9	73.113 \\
36	72.01 \\
36.1	72.152 \\
36.2	70.066 \\
36.3	70.695 \\
36.4	71.212 \\
36.5	72.648 \\
36.6	75.766 \\
36.7	74.491 \\
36.8	76.027 \\
36.9	76.491 \\
37	74.251 \\
37.1	74.468 \\
37.2	75.941 \\
37.3	87.615 \\
37.4	77.03 \\
37.5	79.171 \\
37.6	76.857 \\
37.7	80.594 \\
37.8	80.361 \\
37.9	81.442 \\
38	81.64 \\
38.1	82.39 \\
38.2	81.014 \\
38.3	83.8 \\
38.4	77.26 \\
38.5	81.102 \\
38.6	76.902 \\
38.7	80.503 \\
38.8	78.262 \\
38.9	78.3 \\
39	75.5 \\
39.1	78.25 \\
39.2	75.091 \\
39.3	78.791 \\
39.4	73.519 \\
39.5	77.239 \\
39.6	75.44 \\
39.7	75.753 \\
39.8	74.863 \\
39.9	72.018 \\
40	73.661 \\
40.1	71.417 \\
40.2	72.267 \\
40.3	74.893 \\
40.4	71.679 \\
40.5	72.746 \\
40.6	72.955 \\
40.7	70.908 \\
40.8	71.832 \\
40.9	73.653 \\
41	72.286 \\
41.1	71.653 \\
41.2	72.047 \\
41.3	72.544 \\
41.4	73.333 \\
41.5	71.213 \\
41.6	72.111 \\
41.7	76.626 \\
41.8	73.631 \\
41.9	72.601 \\
42	74.264 \\
42.1	72.918 \\
42.2	77.508 \\
42.3	71.336 \\
42.4	72.469 \\
42.5	75.84 \\
42.6	73.439 \\
42.7	78.092 \\
42.8	71.246 \\
42.9	70.926 \\
43	73.61 \\
43.1	70.446 \\
43.2	73.768 \\
43.3	72.942 \\
43.4	70.573 \\
43.5	71.364 \\
43.6	71.544 \\
43.7	70.449 \\
43.8	70.132 \\
43.9	72.299 \\
44	71.31 \\
44.1	71.769 \\
44.2	72.477 \\
44.3	73.517 \\
44.4	70.448 \\
44.5	70.816 \\
44.6	72.391 \\
44.7	70.339 \\
44.8	69.21 \\
44.9	72.193 \\
45	72.193 \\
45.1	74.399 \\
45.2	73.006 \\
45.3	72.483 \\
45.4	77.537 \\
45.5	70.826 \\
45.6	74.559 \\
45.7	76.167 \\
45.8	76.581 \\
45.9	78.776 \\
46	74.637 \\
46.1	85.26 \\
46.2	77.789 \\
46.3	75.02 \\
46.4	77.876 \\
46.5	75.75 \\
46.6	75.452 \\
46.7	72.965 \\
46.8	76.408 \\
46.9	77.263 \\
47	71.77 \\
47.1	76.448 \\
47.2	72.694 \\
47.3	72.074 \\
47.4	73.278 \\
47.5	72.59 \\
47.6	75.02 \\
47.7	72.179 \\
47.8	71.135 \\
47.9	73.046 \\
48	72.925 \\
48.1	75.813 \\
48.2	77.592 \\
48.3	74.803 \\
48.4	75.077 \\
48.5	72.612 \\
48.6	75.417 \\
48.7	75.795 \\
48.8	73.716 \\
48.9	74.792 \\
49	74.063 \\
49.1	77.572 \\
49.2	83.16 \\
49.3	92.708 \\
49.4	75.65 \\
49.5	72.448 \\
49.6	72.79 \\
49.7	76.928 \\
49.8	75.45 \\
49.9	82.846 \\
50	77.803 \\
50.1	72.64 \\
50.2	78.758 \\
50.3	73.854 \\
50.4	73.774 \\
50.5	80.816 \\
50.6	78.878 \\
50.7	78.792 \\
50.8	80.167 \\
50.9	74.824 \\
51	83.972 \\
51.1	79.729 \\
51.2	83.561 \\
51.3	79.815 \\
51.4	81.338 \\
51.5	86.82 \\
51.6	82.594 \\
51.7	78.789 \\
51.8	81.984 \\
51.9	77.499 \\
52	79.299 \\
52.1	78.242 \\
52.2	85.45 \\
52.3	88.246 \\
52.4	88.886 \\
52.5	82.948 \\
52.6	81.994 \\
52.7	84.904 \\
52.8	85.716 \\
52.9	84.269 \\
53	81.332 \\
53.1	77.362 \\
53.2	77.642 \\
53.3	79.106 \\
53.4	76.73 \\
53.5	80.534 \\
53.6	82.287 \\
53.7	83.176 \\
53.8	82.529 \\
53.9	78.483 \\
54	77.466 \\
54.1	74.056 \\
54.2	71.203 \\
54.3	74.83 \\
54.4	73.436 \\
54.5	78.288 \\
54.6	74.107 \\
54.7	75.577 \\
54.8	77.075 \\
54.9	80.825 \\
55	75.82 \\
55.1	71.924 \\
55.2	77.653 \\
55.3	84.632 \\
55.4	78.398 \\
55.5	79.597 \\
55.6	78.07 \\
55.7	79.613 \\
55.8	77.889 \\
55.9	78.664 \\
56	77.534 \\
56.1	73.888 \\
56.2	74.138 \\
56.3	73.29 \\
56.4	77.107 \\
56.5	71.468 \\
56.6	79.871 \\
56.7	77.677 \\
56.8	79.133 \\
56.9	77.635 \\
57	79.255 \\
57.1	76.344 \\
57.2	76.852 \\
57.3	77.722 \\
57.4	80.626 \\
57.5	74.996 \\
57.6	75.371 \\
57.7	73.839 \\
57.8	75.571 \\
57.9	73.406 \\
58	71.796 \\
58.1	77.334 \\
58.2	72.974 \\
58.3	75.496 \\
58.4	73.56 \\
58.5	75.666 \\
58.6	77.151 \\
58.7	72.992 \\
58.8	73.057 \\
58.9	74.057 \\
59	74.108 \\
59.1	74.873 \\
59.2	71.505 \\
59.3	72.251 \\
59.4	75.173 \\
59.5	73.411 \\
59.6	77.153 \\
59.7	73.071 \\
59.8	74.307 \\
59.9	76.067 \\
60	73.293 \\
60.1	75.241 \\
60.2	77.772 \\
60.3	79.967 \\
60.4	93.711 \\
60.5	73.234 \\
60.6	76.351 \\
60.7	73.508 \\
60.8	71.082 \\
60.9	82.591 \\
61	75.894 \\
61.1	77.37 \\
61.2	73.369 \\
61.3	77.009 \\
61.4	80.343 \\
61.5	73.161 \\
61.6	76.653 \\
61.7	76.585 \\
61.8	74.587 \\
61.9	70.588 \\
62	73.214 \\
62.1	73.048 \\
62.2	74.918 \\
62.3	75.132 \\
62.4	73.517 \\
62.5	79.388 \\
62.6	73.981 \\
62.7	79.684 \\
62.8	71.864 \\
62.9	78.08 \\
63	75.096 \\
63.1	79.746 \\
63.2	76.765 \\
63.3	77.842 \\
63.4	76.669 \\
63.5	81.503 \\
63.6	83.923 \\
63.7	80.152 \\
63.8	85.814 \\
63.9	77.12 \\
64	82.89 \\
64.1	83.62 \\
64.2	81.015 \\
64.3	77.316 \\
64.4	82.038 \\
64.5	82.557 \\
64.6	82.606 \\
64.7	78.585 \\
64.8	79.63 \\
64.9	78.734 \\
65	77.369 \\
65.1	79.719 \\
65.2	76.292 \\
65.3	77.291 \\
65.4	75.351 \\
65.5	77.41 \\
65.6	74.119 \\
65.7	73.864 \\
65.8	74.386 \\
65.9	73.309 \\
66	79.569 \\
66.1	76.157 \\
66.2	81.695 \\
66.3	77.52 \\
66.4	75.605 \\
66.5	73.949 \\
66.6	74.217 \\
66.7	72.695 \\
66.8	75.4 \\
66.9	77.757 \\
67	72.751 \\
67.1	78.432 \\
67.2	73.523 \\
67.3	74.584 \\
67.4	72.858 \\
67.5	78.498 \\
67.6	73.39 \\
67.7	72.292 \\
67.8	75.472 \\
67.9	94.447 \\
68	75.322 \\
68.1	78.7 \\
68.2	71.5 \\
68.3	73.83 \\
68.4	74.489 \\
68.5	75.046 \\
68.6	74.706 \\
68.7	72.786 \\
68.8	73.167 \\
68.9	75.919 \\
69	72.476 \\
69.1	77.561 \\
69.2	76.949 \\
69.3	77.801 \\
69.4	73.641 \\
69.5	75.965 \\
69.6	73.813 \\
69.7	72.94 \\
69.8	76.204 \\
69.9	78.396 \\
70	77.824 \\
70.1	77.72 \\
70.2	79.266 \\
70.3	73.993 \\
70.4	73.988 \\
70.5	72.934 \\
70.6	74.755 \\
70.7	76.947 \\
70.8	78.657 \\
70.9	76.922 \\
71	79.596 \\
71.1	78.582 \\
71.2	81.319 \\
71.3	80.879 \\
71.4	76.744 \\
71.5	75.658 \\
71.6	74.982 \\
71.7	73.022 \\
71.8	71.746 \\
71.9	76.349 \\
72	74.682 \\
72.1	78.288 \\
72.2	77.989 \\
72.3	75.692 \\
72.4	75.544 \\
72.5	76.103 \\
72.6	76.253 \\
72.7	78.37 \\
72.8	73.279 \\
72.9	73.068 \\
73	76.993 \\
73.1	72.941 \\
73.2	72.954 \\
73.3	74.191 \\
73.4	74.327 \\
73.5	75.191 \\
73.6	71.459 \\
73.7	72.488 \\
73.8	77.648 \\
73.9	74.492 \\
74	75.042 \\
74.1	73.786 \\
74.2	76.733 \\
74.3	73.551 \\
74.4	71.359 \\
74.5	69.886 \\
74.6	72.941 \\
74.7	73.522 \\
74.8	72.694 \\
74.9	71.359 \\
75	73.391 \\
75.1	76.48 \\
75.2	75.075 \\
75.3	77.024 \\
75.4	73.793 \\
75.5	73.065 \\
75.6	72.203 \\
75.7	72.51 \\
75.8	72.077 \\
75.9	77.315 \\
76	75.541 \\
76.1	75.031 \\
76.2	72.8 \\
76.3	72.204 \\
76.4	82.253 \\
76.5	77.044 \\
76.6	79.321 \\
76.7	72.317 \\
76.8	74.934 \\
76.9	75.406 \\
77	80.679 \\
77.1	78.631 \\
77.2	83.156 \\
77.3	73.722 \\
77.4	76.657 \\
77.5	76.679 \\
77.6	79.552 \\
77.7	77.925 \\
77.8	80.201 \\
77.9	76.871 \\
78	78.642 \\
78.1	79.607 \\
78.2	75.143 \\
78.3	72.869 \\
78.4	76.656 \\
78.5	75.014 \\
78.6	80.339 \\
};
\addplot [red!89.80392156862746!black]
table [row sep=\\]{%
0	0.086 \\
0.1	0.046 \\
0.2	0.047 \\
0.3	0.033 \\
0.4	0.025 \\
0.5	0.027 \\
0.6	0.048 \\
0.7	0.086 \\
0.8	0.125 \\
0.9	0.046 \\
1	0.047 \\
1.1	0.033 \\
1.2	0.025 \\
1.3	0.027 \\
1.4	0.048 \\
1.5	0.03 \\
1.6	0.002 \\
1.7	0.028 \\
1.8	0.065 \\
1.9	0.002 \\
2	0.047 \\
2.1	0.046 \\
2.2	0.019 \\
2.3	0.029 \\
2.4	0.096 \\
2.5	0.064 \\
2.6	0.064 \\
2.7	0.028 \\
2.8	0.028 \\
2.9	0.06 \\
3	0.048 \\
3.1	0.062 \\
3.2	0.065 \\
3.3	0.05 \\
3.4	0.055 \\
3.5	0.058 \\
3.6	0.033 \\
3.7	0.064 \\
3.8	0.073 \\
3.9	0.067 \\
4	0.064 \\
4.1	0.049 \\
4.2	0.108 \\
4.3	0.048 \\
4.4	0.027 \\
4.5	0.055 \\
4.6	0.049 \\
4.7	0.046 \\
4.8	0.038 \\
4.9	0.03 \\
5	0.068 \\
5.1	0.049 \\
5.2	0.048 \\
5.3	0.032 \\
5.4	0.03 \\
5.5	0.036 \\
5.6	0.068 \\
5.7	0.053 \\
5.8	0.063 \\
5.9	0.052 \\
6	0.03 \\
6.1	0.031 \\
6.2	0.038 \\
6.3	0.064 \\
6.4	0.075 \\
6.5	0.037 \\
6.6	0.134 \\
6.7	0.033 \\
6.8	0.068 \\
6.9	0.079 \\
7	0.07 \\
7.1	0.076 \\
7.2	0.074 \\
7.3	0.057 \\
7.4	0.094 \\
7.5	0.052 \\
7.6	0.002 \\
7.7	0.042 \\
7.8	0.096 \\
7.9	0.071 \\
8	0.09 \\
8.1	0.102 \\
8.2	0.081 \\
8.3	0.056 \\
8.4	0.048 \\
8.5	0.039 \\
8.6	0.057 \\
8.7	0.063 \\
8.8	0.126 \\
8.9	0.085 \\
9	0.052 \\
9.1	0.074 \\
9.2	0.084 \\
9.3	0.257 \\
9.4	0.106 \\
9.5	0.107 \\
9.6	0.079 \\
9.7	0.133 \\
9.8	0.116 \\
9.9	0.105 \\
10	0.074 \\
10.1	0.102 \\
10.2	0.198 \\
10.3	0.138 \\
10.4	0.139 \\
10.5	0.101 \\
10.6	0.153 \\
10.7	0.129 \\
10.8	0.107 \\
10.9	0.112 \\
11	0.158 \\
11.1	0.139 \\
11.2	0.115 \\
11.3	0.186 \\
11.4	0.191 \\
11.5	0.151 \\
11.6	0.177 \\
11.7	0.221 \\
11.8	0.292 \\
11.9	0.204 \\
12	0.202 \\
12.1	0.274 \\
12.2	0.241 \\
12.3	0.535 \\
12.4	0.239 \\
12.5	0.235 \\
12.6	0.276 \\
12.7	0.312 \\
12.8	0.251 \\
12.9	0.55 \\
13	0.339 \\
13.1	0.506 \\
13.2	0.374 \\
13.3	0.566 \\
13.4	0.358 \\
13.5	0.399 \\
13.6	0.412 \\
13.7	0.566 \\
13.8	0.499 \\
13.9	1.182 \\
14	0.418 \\
14.1	0.477 \\
14.2	0.415 \\
14.3	0.415 \\
14.4	0.806 \\
14.5	0.48 \\
14.6	0.829 \\
14.7	0.484 \\
14.8	0.489 \\
14.9	0.428 \\
15	0.696 \\
15.1	1.126 \\
15.2	0.574 \\
15.3	0.427 \\
15.4	0.451 \\
15.5	0.767 \\
15.6	0.44 \\
15.7	0.444 \\
15.8	0.607 \\
15.9	0.557 \\
16	0.594 \\
16.1	0.546 \\
16.2	0.51 \\
16.3	0.492 \\
16.4	1.519 \\
16.5	0.511 \\
16.6	0.482 \\
16.7	0.754 \\
16.8	0.515 \\
16.9	0.913 \\
17	0.55 \\
17.1	0.649 \\
17.2	0.493 \\
17.3	0.002 \\
17.4	0.029 \\
17.5	0.002 \\
17.6	0.003 \\
17.7	0.002 \\
17.8	0.003 \\
17.9	0.023 \\
18	0.003 \\
18.1	0.002 \\
18.2	0.019 \\
18.3	0.002 \\
18.4	0.021 \\
18.5	0.022 \\
18.6	0.006 \\
18.7	0.002 \\
18.8	0.002 \\
18.9	0.003 \\
19	0.003 \\
19.1	0.002 \\
19.2	0.202 \\
19.3	0.02 \\
19.4	0.002 \\
19.5	0.023 \\
19.6	0.003 \\
19.7	0.002 \\
19.8	0.003 \\
19.9	0.007 \\
20	0.002 \\
20.1	0.032 \\
20.2	0.003 \\
20.3	0.002 \\
20.4	0.004 \\
20.5	0.002 \\
20.6	0.003 \\
20.7	0.003 \\
20.8	0.003 \\
20.9	0.003 \\
21	0.003 \\
21.1	0.029 \\
21.2	0.002 \\
21.3	0.003 \\
21.4	0.003 \\
21.5	0.021 \\
21.6	0.002 \\
21.7	0.003 \\
21.8	0.004 \\
21.9	0.002 \\
22	0.002 \\
22.1	0.004 \\
22.2	0.002 \\
22.3	0.002 \\
22.4	0.015 \\
22.5	0.037 \\
22.6	0.002 \\
22.7	0.003 \\
22.8	0.002 \\
22.9	0.003 \\
23	0.019 \\
23.1	0.002 \\
23.2	0.003 \\
23.3	0.063 \\
23.4	0.042 \\
23.5	0.044 \\
23.6	0.044 \\
23.7	0.062 \\
23.8	0.067 \\
23.9	0.102 \\
24	0.072 \\
24.1	0.088 \\
24.2	0.074 \\
24.3	0.052 \\
24.4	0.056 \\
24.5	0.123 \\
24.6	0.083 \\
24.7	0.125 \\
24.8	0.051 \\
24.9	0.073 \\
25	0.089 \\
25.1	0.123 \\
25.2	0.082 \\
25.3	0.128 \\
25.4	0.108 \\
25.5	0.113 \\
25.6	0.146 \\
25.7	0.077 \\
25.8	0.138 \\
25.9	0.174 \\
26	0.14 \\
26.1	0.12 \\
26.2	0.139 \\
26.3	0.08 \\
26.4	0.231 \\
26.5	0.138 \\
26.6	0.168 \\
26.7	0.1 \\
26.8	0.091 \\
26.9	0.163 \\
27	0.08 \\
27.1	0.109 \\
27.2	0.094 \\
27.3	0.069 \\
27.4	0.182 \\
27.5	0.108 \\
27.6	0.137 \\
27.7	0.202 \\
27.8	0.119 \\
27.9	0.093 \\
28	0.076 \\
28.1	0.113 \\
28.2	0.129 \\
28.3	0.142 \\
28.4	0.12 \\
28.5	0.137 \\
28.6	0.095 \\
28.7	0.083 \\
28.8	0.14 \\
28.9	0.073 \\
29	0.094 \\
29.1	0.101 \\
29.2	0.28 \\
29.3	0.102 \\
29.4	0.173 \\
29.5	0.134 \\
29.6	0.089 \\
29.7	0.084 \\
29.8	0.122 \\
29.9	0.097 \\
30	0.109 \\
30.1	0.093 \\
30.2	0.098 \\
30.3	0.115 \\
30.4	0.147 \\
30.5	0.155 \\
30.6	0.117 \\
30.7	0.075 \\
30.8	0.091 \\
30.9	0.089 \\
31	0.068 \\
31.1	0.069 \\
31.2	0.086 \\
31.3	0.096 \\
31.4	0.081 \\
31.5	0.003 \\
31.6	0.125 \\
31.7	0.003 \\
31.8	0.099 \\
31.9	0.202 \\
32	0.106 \\
32.1	0.107 \\
32.2	0.154 \\
32.3	0.136 \\
32.4	0.092 \\
32.5	0.108 \\
32.6	0.252 \\
32.7	0.13 \\
32.8	0.096 \\
32.9	0.126 \\
33	0.09 \\
33.1	0.103 \\
33.2	0.1 \\
33.3	0.106 \\
33.4	0.097 \\
33.5	0.104 \\
33.6	0.096 \\
33.7	0.115 \\
33.8	0.1 \\
33.9	0.065 \\
34	0.081 \\
34.1	0.098 \\
34.2	0.079 \\
34.3	0.088 \\
34.4	0.106 \\
34.5	0.074 \\
34.6	0.092 \\
34.7	0.103 \\
34.8	0.15 \\
34.9	0.104 \\
35	0.003 \\
35.1	0.003 \\
35.2	0.005 \\
35.3	0.002 \\
35.4	0.021 \\
35.5	0.038 \\
35.6	0.019 \\
35.7	0.021 \\
35.8	0.002 \\
35.9	0.002 \\
36	0.002 \\
36.1	0.242 \\
36.2	0.075 \\
36.3	0.089 \\
36.4	0.086 \\
36.5	0.092 \\
36.6	0.088 \\
36.7	0.281 \\
36.8	0.098 \\
36.9	0.133 \\
37	0.085 \\
37.1	0.088 \\
37.2	0.13 \\
37.3	0.292 \\
37.4	0.128 \\
37.5	0.12 \\
37.6	0.216 \\
37.7	0.142 \\
37.8	0.162 \\
37.9	0.213 \\
38	0.232 \\
38.1	0.429 \\
38.2	0.305 \\
38.3	0.472 \\
38.4	0.334 \\
38.5	0.273 \\
38.6	0.21 \\
38.7	0.232 \\
38.8	0.334 \\
38.9	0.228 \\
39	0.201 \\
39.1	0.286 \\
39.2	0.346 \\
39.3	0.258 \\
39.4	0.263 \\
39.5	0.28 \\
39.6	0.276 \\
39.7	0.313 \\
39.8	0.266 \\
39.9	0.244 \\
40	0.003 \\
40.1	0.002 \\
40.2	0.021 \\
40.3	0.002 \\
40.4	0.249 \\
40.5	0.435 \\
40.6	0.332 \\
40.7	0.002 \\
40.8	0.004 \\
40.9	0.002 \\
41	0.002 \\
41.1	0.003 \\
41.2	0.284 \\
41.3	0.259 \\
41.4	0.008 \\
41.5	0.003 \\
41.6	0.002 \\
41.7	0.002 \\
41.8	0.003 \\
41.9	0.003 \\
42	0.002 \\
42.1	0.019 \\
42.2	0.004 \\
42.3	0.022 \\
42.4	0.002 \\
42.5	0.019 \\
42.6	0.002 \\
42.7	0.003 \\
42.8	0.002 \\
42.9	0.022 \\
43	0.028 \\
43.1	0.002 \\
43.2	0.002 \\
43.3	0.019 \\
43.4	0.021 \\
43.5	0.019 \\
43.6	0.002 \\
43.7	0.021 \\
43.8	0.002 \\
43.9	0.003 \\
44	0.002 \\
44.1	0.007 \\
44.2	0.002 \\
44.3	0.002 \\
44.4	0.002 \\
44.5	0.009 \\
44.6	0.002 \\
44.7	0.002 \\
44.8	0.002 \\
44.9	0.002 \\
45	0.002 \\
45.1	0.002 \\
45.2	0.002 \\
45.3	0.002 \\
45.4	0.002 \\
45.5	0.003 \\
45.6	0.003 \\
45.7	0.003 \\
45.8	0.003 \\
45.9	0.003 \\
46	0.002 \\
46.1	0.007 \\
46.2	0.029 \\
46.3	0.002 \\
46.4	0.019 \\
46.5	0.003 \\
46.6	0.003 \\
46.7	0.002 \\
46.8	0.019 \\
46.9	0.019 \\
47	0.002 \\
47.1	0.002 \\
47.2	0.021 \\
47.3	0.003 \\
47.4	0.019 \\
47.5	0.02 \\
47.6	0.006 \\
47.7	0.021 \\
47.8	0.002 \\
47.9	0.003 \\
48	0.003 \\
48.1	0.205 \\
48.2	0.048 \\
48.3	0.028 \\
48.4	0.032 \\
48.5	0.039 \\
48.6	0.062 \\
48.7	0.105 \\
48.8	0.053 \\
48.9	0.042 \\
49	0.038 \\
49.1	0.048 \\
49.2	0.155 \\
49.3	0.061 \\
49.4	0.089 \\
49.5	0.082 \\
49.6	0.113 \\
49.7	0.072 \\
49.8	0.131 \\
49.9	0.092 \\
50	0.169 \\
50.1	0.112 \\
50.2	0.09 \\
50.3	0.116 \\
50.4	0.132 \\
50.5	0.163 \\
50.6	0.201 \\
50.7	0.131 \\
50.8	0.174 \\
50.9	0.167 \\
51	0.167 \\
51.1	0.128 \\
51.2	0.193 \\
51.3	0.203 \\
51.4	0.252 \\
51.5	0.206 \\
51.6	0.204 \\
51.7	0.154 \\
51.8	0.206 \\
51.9	0.223 \\
52	0.202 \\
52.1	0.197 \\
52.2	0.166 \\
52.3	0.18 \\
52.4	0.223 \\
52.5	0.791 \\
52.6	0.239 \\
52.7	0.239 \\
52.8	0.392 \\
52.9	0.186 \\
53	0.231 \\
53.1	0.473 \\
53.2	0.248 \\
53.3	0.482 \\
53.4	0.33 \\
53.5	0.657 \\
53.6	0.274 \\
53.7	0.241 \\
53.8	0.441 \\
53.9	0.216 \\
54	0.692 \\
54.1	0.222 \\
54.2	0.292 \\
54.3	0.205 \\
54.4	0.25 \\
54.5	0.381 \\
54.6	0.226 \\
54.7	0.305 \\
54.8	0.254 \\
54.9	0.875 \\
55	0.298 \\
55.1	0.276 \\
55.2	0.354 \\
55.3	0.259 \\
55.4	0.29 \\
55.5	0.303 \\
55.6	0.304 \\
55.7	0.27 \\
55.8	0.294 \\
55.9	0.287 \\
56	0.26 \\
56.1	0.237 \\
56.2	0.28 \\
56.3	0.284 \\
56.4	0.246 \\
56.5	0.328 \\
56.6	0.36 \\
56.7	0.311 \\
56.8	0.752 \\
56.9	0.34 \\
57	0.297 \\
57.1	0.253 \\
57.2	0.387 \\
57.3	0.244 \\
57.4	0.274 \\
57.5	0.268 \\
57.6	0.276 \\
57.7	0.273 \\
57.8	0.344 \\
57.9	0.322 \\
58	0.181 \\
58.1	0.286 \\
58.2	0.194 \\
58.3	1.683 \\
58.4	0.353 \\
58.5	0.326 \\
58.6	0.276 \\
58.7	0.285 \\
58.8	0.505 \\
58.9	0.266 \\
59	0.303 \\
59.1	0.338 \\
59.2	0.151 \\
59.3	0.151 \\
59.4	0.154 \\
59.5	0.179 \\
59.6	0.259 \\
59.7	0.27 \\
59.8	0.146 \\
59.9	0.161 \\
60	0.394 \\
60.1	0.133 \\
60.2	0.307 \\
60.3	0.343 \\
60.4	0.148 \\
60.5	0.427 \\
60.6	0.155 \\
60.7	0.138 \\
60.8	0.116 \\
60.9	0.166 \\
61	0.165 \\
61.1	0.224 \\
61.2	0.156 \\
61.3	0.176 \\
61.4	0.66 \\
61.5	0.256 \\
61.6	0.152 \\
61.7	0.238 \\
61.8	0.059 \\
61.9	0.177 \\
62	0.082 \\
62.1	0.075 \\
62.2	0.167 \\
62.3	0.101 \\
62.4	0.155 \\
62.5	0.197 \\
62.6	0.203 \\
62.7	0.12 \\
62.8	0.122 \\
62.9	0.118 \\
63	0.135 \\
63.1	0.168 \\
63.2	0.209 \\
63.3	0.124 \\
63.4	0.191 \\
63.5	0.168 \\
63.6	0.403 \\
63.7	0.264 \\
63.8	0.291 \\
63.9	0.23 \\
64	0.256 \\
64.1	0.213 \\
64.2	0.28 \\
64.3	0.424 \\
64.4	0.684 \\
64.5	0.248 \\
64.6	0.454 \\
64.7	0.385 \\
64.8	0.25 \\
64.9	0.279 \\
65	0.334 \\
65.1	0.24 \\
65.2	0.258 \\
65.3	0.275 \\
65.4	0.303 \\
65.5	0.307 \\
65.6	0.229 \\
65.7	0.326 \\
65.8	0.277 \\
65.9	0.416 \\
66	0.276 \\
66.1	0.335 \\
66.2	0.307 \\
66.3	0.542 \\
66.4	0.358 \\
66.5	0.306 \\
66.6	0.311 \\
66.7	0.317 \\
66.8	0.627 \\
66.9	0.305 \\
67	0.586 \\
67.1	0.239 \\
67.2	0.317 \\
67.3	0.236 \\
67.4	0.275 \\
67.5	0.265 \\
67.6	0.237 \\
67.7	0.251 \\
67.8	0.266 \\
67.9	0.276 \\
68	0.311 \\
68.1	0.002 \\
68.2	0.003 \\
68.3	0.006 \\
68.4	0.002 \\
68.5	0.002 \\
68.6	0.004 \\
68.7	0.002 \\
68.8	0.004 \\
68.9	0.02 \\
69	0.02 \\
69.1	0.02 \\
69.2	0.003 \\
69.3	0.003 \\
69.4	0.003 \\
69.5	0.008 \\
69.6	0.002 \\
69.7	0.258 \\
69.8	0.053 \\
69.9	0.053 \\
70	0.061 \\
70.1	0.045 \\
70.2	0.093 \\
70.3	0.138 \\
70.4	0.046 \\
70.5	0.048 \\
70.6	0.106 \\
70.7	0.058 \\
70.8	0.071 \\
70.9	0.048 \\
71	0.07 \\
71.1	0.031 \\
71.2	0.068 \\
71.3	0.048 \\
71.4	0.081 \\
71.5	0.028 \\
71.6	0.121 \\
71.7	0.063 \\
71.8	0.03 \\
71.9	0.03 \\
72	0.031 \\
72.1	0.05 \\
72.2	0.098 \\
72.3	0.035 \\
72.4	0.155 \\
72.5	0.03 \\
72.6	0.032 \\
72.7	0.049 \\
72.8	0.03 \\
72.9	0.06 \\
73	0.031 \\
73.1	0.031 \\
73.2	0.034 \\
73.3	0.069 \\
73.4	0.053 \\
73.5	0.033 \\
73.6	0.031 \\
73.7	0.051 \\
73.8	0.032 \\
73.9	0.031 \\
74	0.038 \\
74.1	0.052 \\
74.2	0.116 \\
74.3	0.032 \\
74.4	0.031 \\
74.5	0.07 \\
74.6	0.075 \\
74.7	0.132 \\
74.8	0.101 \\
74.9	0.032 \\
75	0.065 \\
75.1	0.052 \\
75.2	0.069 \\
75.3	0.061 \\
75.4	0.076 \\
75.5	0.022 \\
75.6	0.003 \\
75.7	0.002 \\
75.8	0.023 \\
75.9	0.003 \\
76	0.002 \\
76.1	0.002 \\
76.2	0.002 \\
76.3	0.002 \\
76.4	0.002 \\
76.5	0.004 \\
76.6	0.019 \\
76.7	0.002 \\
76.8	0.002 \\
76.9	0.004 \\
77	0.019 \\
77.1	0.002 \\
77.2	0.022 \\
77.3	0.002 \\
77.4	0.003 \\
77.5	0.002 \\
77.6	0.021 \\
77.7	0.016 \\
77.8	0.002 \\
77.9	0.003 \\
78	0.002 \\
78.1	0.003 \\
78.2	0.003 \\
78.3	0.003 \\
78.4	0.002 \\
78.5	0.002 \\
78.6	0.002 \\
};
\addplot [color0]
table [row sep=\\]{%
0	1.178 \\
0.1	1.563 \\
0.2	1.607 \\
0.3	0.844 \\
0.4	0.852 \\
0.5	0.862 \\
0.6	1.336 \\
0.7	0.953 \\
0.8	0.822 \\
0.9	0.875 \\
1	0.846 \\
1.1	0.862 \\
1.2	0.81 \\
1.3	0.886 \\
1.4	1.211 \\
1.5	0.864 \\
1.6	0.856 \\
1.7	0.79 \\
1.8	0.811 \\
1.9	0.848 \\
2	1.057 \\
2.1	0.866 \\
2.2	0.792 \\
2.3	1.013 \\
2.4	0.962 \\
2.5	1.566 \\
2.6	0.992 \\
2.7	1.171 \\
2.8	0.875 \\
2.9	1.257 \\
3	1.008 \\
3.1	0.934 \\
3.2	0.912 \\
3.3	0.917 \\
3.4	1.227 \\
3.5	0.922 \\
3.6	0.868 \\
3.7	0.84 \\
3.8	1.833 \\
3.9	0.881 \\
4	0.959 \\
4.1	0.866 \\
4.2	1.544 \\
4.3	0.813 \\
4.4	0.8 \\
4.5	1.156 \\
4.6	0.896 \\
4.7	0.857 \\
4.8	1.465 \\
4.9	0.95 \\
5	0.859 \\
5.1	1.056 \\
5.2	0.847 \\
5.3	0.991 \\
5.4	0.829 \\
5.5	1.106 \\
5.6	0.833 \\
5.7	0.847 \\
5.8	1.578 \\
5.9	0.925 \\
6	0.827 \\
6.1	0.826 \\
6.2	1.428 \\
6.3	0.835 \\
6.4	0.805 \\
6.5	0.932 \\
6.6	1.327 \\
6.7	0.879 \\
6.8	1.023 \\
6.9	1.083 \\
7	0.815 \\
7.1	0.89 \\
7.2	0.808 \\
7.3	1.016 \\
7.4	0.911 \\
7.5	0.896 \\
7.6	0.827 \\
7.7	1.136 \\
7.8	1.094 \\
7.9	0.846 \\
8	1.17 \\
8.1	0.911 \\
8.2	0.835 \\
8.3	0.827 \\
8.4	1.392 \\
8.5	0.903 \\
8.6	0.821 \\
8.7	0.932 \\
8.8	1.971 \\
8.9	0.846 \\
9	1.618 \\
9.1	0.844 \\
9.2	1.338 \\
9.3	1.528 \\
9.4	0.845 \\
9.5	1.149 \\
9.6	1.103 \\
9.7	1.092 \\
9.8	0.877 \\
9.9	1.354 \\
10	0.89 \\
10.1	1.091 \\
10.2	1.826 \\
10.3	0.917 \\
10.4	0.91 \\
10.5	1.232 \\
10.6	1.106 \\
10.7	0.992 \\
10.8	1.34 \\
10.9	0.978 \\
11	1.295 \\
11.1	0.963 \\
11.2	0.977 \\
11.3	1.034 \\
11.4	0.922 \\
11.5	1.059 \\
11.6	0.928 \\
11.7	1.045 \\
11.8	1.311 \\
11.9	0.968 \\
12	0.949 \\
12.1	1.616 \\
12.2	1.29 \\
12.3	1.235 \\
12.4	1.138 \\
12.5	1.012 \\
12.6	1.143 \\
12.7	1.092 \\
12.8	1.109 \\
12.9	1.477 \\
13	1.04 \\
13.1	2.023 \\
13.2	1.026 \\
13.3	1.659 \\
13.4	1.167 \\
13.5	1.765 \\
13.6	0.999 \\
13.7	1.619 \\
13.8	0.995 \\
13.9	2.312 \\
14	1.005 \\
14.1	1.151 \\
14.2	0.941 \\
14.3	0.991 \\
14.4	1.189 \\
14.5	1.838 \\
14.6	0.933 \\
14.7	0.956 \\
14.8	1.04 \\
14.9	1.621 \\
15	0.911 \\
15.1	1.868 \\
15.2	0.87 \\
15.3	0.948 \\
15.4	0.921 \\
15.5	1.104 \\
15.6	1.023 \\
15.7	0.894 \\
15.8	1.551 \\
15.9	1.259 \\
16	0.86 \\
16.1	0.871 \\
16.2	0.834 \\
16.3	1.166 \\
16.4	1.77 \\
16.5	0.933 \\
16.6	0.913 \\
16.7	1.551 \\
16.8	0.936 \\
16.9	0.864 \\
17	0.868 \\
17.1	0.909 \\
17.2	0.84 \\
17.3	0.902 \\
17.4	0.803 \\
17.5	0.832 \\
17.6	0.827 \\
17.7	0.847 \\
17.8	0.826 \\
17.9	0.814 \\
18	1.223 \\
18.1	0.904 \\
18.2	0.858 \\
18.3	0.813 \\
18.4	1.123 \\
18.5	0.885 \\
18.6	1.415 \\
18.7	0.831 \\
18.8	0.797 \\
18.9	1.067 \\
19	1.685 \\
19.1	0.854 \\
19.2	0.97 \\
19.3	0.899 \\
19.4	0.883 \\
19.5	1.779 \\
19.6	0.801 \\
19.7	0.798 \\
19.8	0.947 \\
19.9	0.872 \\
20	1.067 \\
20.1	0.914 \\
20.2	0.964 \\
20.3	0.829 \\
20.4	1.511 \\
20.5	0.825 \\
20.6	0.898 \\
20.7	0.792 \\
20.8	0.799 \\
20.9	0.773 \\
21	1.714 \\
21.1	0.868 \\
21.2	0.879 \\
21.3	0.872 \\
21.4	0.814 \\
21.5	0.842 \\
21.6	0.947 \\
21.7	1.35 \\
21.8	1.758 \\
21.9	0.814 \\
22	0.882 \\
22.1	1.863 \\
22.2	0.828 \\
22.3	0.856 \\
22.4	0.87 \\
22.5	1.677 \\
22.6	0.899 \\
22.7	1.042 \\
22.8	0.922 \\
22.9	0.955 \\
23	0.838 \\
23.1	0.9 \\
23.2	0.909 \\
23.3	1.082 \\
23.4	0.86 \\
23.5	0.805 \\
23.6	1.312 \\
23.7	0.859 \\
23.8	0.829 \\
23.9	1.821 \\
24	0.919 \\
24.1	0.925 \\
24.2	0.926 \\
24.3	0.889 \\
24.4	0.957 \\
24.5	1.285 \\
24.6	1.059 \\
24.7	1.886 \\
24.8	0.958 \\
24.9	0.984 \\
25	0.896 \\
25.1	0.931 \\
25.2	0.875 \\
25.3	1.241 \\
25.4	0.966 \\
25.5	0.873 \\
25.6	1.388 \\
25.7	0.855 \\
25.8	0.864 \\
25.9	0.942 \\
26	0.895 \\
26.1	0.915 \\
26.2	0.882 \\
26.3	0.861 \\
26.4	1.816 \\
26.5	0.994 \\
26.6	1.78 \\
26.7	0.9 \\
26.8	1.049 \\
26.9	1.695 \\
27	0.942 \\
27.1	0.947 \\
27.2	0.917 \\
27.3	0.89 \\
27.4	1.075 \\
27.5	0.937 \\
27.6	0.835 \\
27.7	0.812 \\
27.8	0.807 \\
27.9	0.924 \\
28	0.903 \\
28.1	0.887 \\
28.2	0.913 \\
28.3	0.846 \\
28.4	0.998 \\
28.5	0.774 \\
28.6	0.883 \\
28.7	0.842 \\
28.8	0.816 \\
28.9	1.518 \\
29	1.136 \\
29.1	0.885 \\
29.2	1.744 \\
29.3	0.894 \\
29.4	1.785 \\
29.5	0.816 \\
29.6	0.866 \\
29.7	0.909 \\
29.8	0.975 \\
29.9	0.915 \\
30	0.866 \\
30.1	0.864 \\
30.2	1.115 \\
30.3	0.816 \\
30.4	0.883 \\
30.5	1.021 \\
30.6	0.841 \\
30.7	0.937 \\
30.8	0.82 \\
30.9	0.831 \\
31	0.908 \\
31.1	1.377 \\
31.2	0.828 \\
31.3	0.887 \\
31.4	0.789 \\
31.5	0.884 \\
31.6	1.424 \\
31.7	0.893 \\
31.8	0.824 \\
31.9	1.643 \\
32	0.791 \\
32.1	0.811 \\
32.2	0.946 \\
32.3	1.309 \\
32.4	0.833 \\
32.5	0.876 \\
32.6	1.953 \\
32.7	0.977 \\
32.8	0.852 \\
32.9	0.831 \\
33	1.146 \\
33.1	0.941 \\
33.2	1.04 \\
33.3	0.99 \\
33.4	0.819 \\
33.5	0.946 \\
33.6	0.849 \\
33.7	0.82 \\
33.8	0.862 \\
33.9	0.885 \\
34	1.155 \\
34.1	0.883 \\
34.2	0.829 \\
34.3	0.84 \\
34.4	1.183 \\
34.5	0.85 \\
34.6	0.825 \\
34.7	1.06 \\
34.8	0.848 \\
34.9	0.96 \\
35	0.876 \\
35.1	0.836 \\
35.2	1.844 \\
35.3	0.937 \\
35.4	0.804 \\
35.5	1.186 \\
35.6	0.851 \\
35.7	1.004 \\
35.8	1.806 \\
35.9	0.867 \\
36	0.991 \\
36.1	1.46 \\
36.2	0.887 \\
36.3	1.059 \\
36.4	1.165 \\
36.5	1.008 \\
36.6	1.269 \\
36.7	1.981 \\
36.8	0.953 \\
36.9	1.255 \\
37	1.028 \\
37.1	1.011 \\
37.2	1.128 \\
37.3	1.947 \\
37.4	1.134 \\
37.5	1.916 \\
37.6	1.095 \\
37.7	1.11 \\
37.8	1.208 \\
37.9	1.043 \\
38	1.827 \\
38.1	1.411 \\
38.2	1.227 \\
38.3	3.096 \\
38.4	1.246 \\
38.5	1.585 \\
38.6	1.16 \\
38.7	1.388 \\
38.8	1.152 \\
38.9	1.101 \\
39	1.062 \\
39.1	1.073 \\
39.2	1.056 \\
39.3	1.06 \\
39.4	0.911 \\
39.5	1.293 \\
39.6	1.055 \\
39.7	1.477 \\
39.8	0.97 \\
39.9	0.893 \\
40	1.154 \\
40.1	0.967 \\
40.2	0.883 \\
40.3	0.88 \\
40.4	0.878 \\
40.5	1.518 \\
40.6	0.816 \\
40.7	0.866 \\
40.8	1.318 \\
40.9	0.811 \\
41	0.834 \\
41.1	1.145 \\
41.2	0.871 \\
41.3	0.928 \\
41.4	1.45 \\
41.5	0.87 \\
41.6	0.82 \\
41.7	1.184 \\
41.8	1.359 \\
41.9	1.941 \\
42	0.842 \\
42.1	0.886 \\
42.2	0.839 \\
42.3	0.919 \\
42.4	1.204 \\
42.5	0.798 \\
42.6	0.835 \\
42.7	0.797 \\
42.8	0.838 \\
42.9	0.879 \\
43	1.179 \\
43.1	0.874 \\
43.2	0.764 \\
43.3	0.804 \\
43.4	0.831 \\
43.5	0.817 \\
43.6	0.938 \\
43.7	0.831 \\
43.8	0.869 \\
43.9	1.531 \\
44	0.843 \\
44.1	0.819 \\
44.2	1.201 \\
44.3	0.787 \\
44.4	0.795 \\
44.5	1.424 \\
44.6	0.788 \\
44.7	0.799 \\
44.8	0.869 \\
44.9	0.834 \\
45	0.924 \\
45.1	2.023 \\
45.2	0.954 \\
45.3	0.787 \\
45.4	0.823 \\
45.5	0.862 \\
45.6	1.666 \\
45.7	0.888 \\
45.8	0.79 \\
45.9	0.97 \\
46	1.007 \\
46.1	1.454 \\
46.2	1.041 \\
46.3	0.803 \\
46.4	0.924 \\
46.5	1.036 \\
46.6	1.313 \\
46.7	0.803 \\
46.8	1.061 \\
46.9	0.965 \\
47	0.821 \\
47.1	1.449 \\
47.2	0.932 \\
47.3	1.018 \\
47.4	0.923 \\
47.5	0.857 \\
47.6	1.859 \\
47.7	0.969 \\
47.8	0.927 \\
47.9	1.015 \\
48	1.012 \\
48.1	0.837 \\
48.2	1.42 \\
48.3	0.889 \\
48.4	0.894 \\
48.5	0.984 \\
48.6	1.168 \\
48.7	0.863 \\
48.8	0.843 \\
48.9	0.889 \\
49	0.851 \\
49.1	2.016 \\
49.2	2.028 \\
49.3	1.138 \\
49.4	0.872 \\
49.5	0.888 \\
49.6	1.326 \\
49.7	0.998 \\
49.8	1.418 \\
49.9	0.956 \\
50	1.974 \\
50.1	0.958 \\
50.2	1.788 \\
50.3	1.113 \\
50.4	0.945 \\
50.5	0.999 \\
50.6	1.637 \\
50.7	0.969 \\
50.8	2.191 \\
50.9	0.902 \\
51	0.982 \\
51.1	1.018 \\
51.2	0.926 \\
51.3	1.074 \\
51.4	1.219 \\
51.5	0.962 \\
51.6	0.991 \\
51.7	1.541 \\
51.8	1.079 \\
51.9	0.981 \\
52	1.063 \\
52.1	1.116 \\
52.2	0.998 \\
52.3	1.115 \\
52.4	1.086 \\
52.5	2.042 \\
52.6	1.483 \\
52.7	0.977 \\
52.8	1.666 \\
52.9	1.05 \\
53	1.094 \\
53.1	1.511 \\
53.2	1.016 \\
53.3	2.455 \\
53.4	1.184 \\
53.5	1.536 \\
53.6	1.618 \\
53.7	1.048 \\
53.8	2.096 \\
53.9	0.936 \\
54	1.671 \\
54.1	0.971 \\
54.2	0.997 \\
54.3	1.092 \\
54.4	1.157 \\
54.5	1.341 \\
54.6	0.904 \\
54.7	1.661 \\
54.8	1.056 \\
54.9	1.159 \\
55	0.893 \\
55.1	0.965 \\
55.2	1.144 \\
55.3	1.325 \\
55.4	1.022 \\
55.5	1.035 \\
55.6	1.107 \\
55.7	0.893 \\
55.8	1 \\
55.9	0.982 \\
56	0.867 \\
56.1	1.177 \\
56.2	0.864 \\
56.3	0.843 \\
56.4	0.913 \\
56.5	1.016 \\
56.6	1.328 \\
56.7	1.037 \\
56.8	2.851 \\
56.9	0.889 \\
57	1.153 \\
57.1	0.84 \\
57.2	1.172 \\
57.3	0.882 \\
57.4	0.868 \\
57.5	0.834 \\
57.6	1.104 \\
57.7	1.023 \\
57.8	1.991 \\
57.9	1.179 \\
58	0.896 \\
58.1	1.298 \\
58.2	0.837 \\
58.3	1.828 \\
58.4	0.988 \\
58.5	0.877 \\
58.6	0.866 \\
58.7	0.832 \\
58.8	1.451 \\
58.9	0.827 \\
59	0.838 \\
59.1	0.935 \\
59.2	0.961 \\
59.3	0.987 \\
59.4	0.826 \\
59.5	1 \\
59.6	1.655 \\
59.7	0.896 \\
59.8	1.025 \\
59.9	0.982 \\
60	0.867 \\
60.1	1.328 \\
60.2	0.827 \\
60.3	0.903 \\
60.4	1.008 \\
60.5	1.576 \\
60.6	1.316 \\
60.7	0.853 \\
60.8	0.882 \\
60.9	1.055 \\
61	0.901 \\
61.1	0.947 \\
61.2	0.899 \\
61.3	0.914 \\
61.4	1.295 \\
61.5	0.921 \\
61.6	0.884 \\
61.7	1.012 \\
61.8	0.863 \\
61.9	0.952 \\
62	0.859 \\
62.1	0.888 \\
62.2	1.173 \\
62.3	0.896 \\
62.4	1.193 \\
62.5	1.046 \\
62.6	1.07 \\
62.7	0.939 \\
62.8	0.96 \\
62.9	0.934 \\
63	0.952 \\
63.1	1.091 \\
63.2	1.069 \\
63.3	1.101 \\
63.4	1.114 \\
63.5	1.114 \\
63.6	2.18 \\
63.7	2.083 \\
63.8	1.437 \\
63.9	1.518 \\
64	1.189 \\
64.1	1.15 \\
64.2	1.388 \\
64.3	1.099 \\
64.4	2.367 \\
64.5	1.019 \\
64.6	1.365 \\
64.7	1.452 \\
64.8	1.277 \\
64.9	1.364 \\
65	1.199 \\
65.1	0.997 \\
65.2	1.008 \\
65.3	0.993 \\
65.4	1.042 \\
65.5	1.035 \\
65.6	0.954 \\
65.7	1.198 \\
65.8	0.972 \\
65.9	1.404 \\
66	0.901 \\
66.1	1.119 \\
66.2	0.929 \\
66.3	1.625 \\
66.4	1.134 \\
66.5	1.281 \\
66.6	0.915 \\
66.7	0.907 \\
66.8	1.379 \\
66.9	0.875 \\
67	0.832 \\
67.1	0.949 \\
67.2	0.847 \\
67.3	1.431 \\
67.4	0.927 \\
67.5	1.112 \\
67.6	0.861 \\
67.7	0.936 \\
67.8	0.836 \\
67.9	0.845 \\
68	1 \\
68.1	0.896 \\
68.2	0.93 \\
68.3	2.263 \\
68.4	0.821 \\
68.5	0.835 \\
68.6	1.439 \\
68.7	0.855 \\
68.8	1.629 \\
68.9	1.007 \\
69	0.899 \\
69.1	0.836 \\
69.2	1.073 \\
69.3	0.913 \\
69.4	0.81 \\
69.5	1.71 \\
69.6	0.896 \\
69.7	0.92 \\
69.8	1.137 \\
69.9	0.802 \\
70	0.79 \\
70.1	0.924 \\
70.2	0.85 \\
70.3	1.392 \\
70.4	1.493 \\
70.5	0.8 \\
70.6	0.959 \\
70.7	0.803 \\
70.8	2.081 \\
70.9	0.913 \\
71	1.738 \\
71.1	0.97 \\
71.2	1.999 \\
71.3	0.881 \\
71.4	0.891 \\
71.5	0.827 \\
71.6	1.43 \\
71.7	0.853 \\
71.8	0.828 \\
71.9	0.863 \\
72	0.849 \\
72.1	1.65 \\
72.2	1.789 \\
72.3	0.957 \\
72.4	1.259 \\
72.5	1.215 \\
72.6	0.968 \\
72.7	1.391 \\
72.8	0.877 \\
72.9	0.85 \\
73	0.842 \\
73.1	0.863 \\
73.2	1.6 \\
73.3	0.839 \\
73.4	0.81 \\
73.5	1.471 \\
73.6	0.994 \\
73.7	0.901 \\
73.8	0.955 \\
73.9	0.835 \\
74	1.088 \\
74.1	0.946 \\
74.2	1.302 \\
74.3	0.814 \\
74.4	0.851 \\
74.5	1.03 \\
74.6	0.892 \\
74.7	1.666 \\
74.8	1.625 \\
74.9	0.831 \\
75	0.828 \\
75.1	1.69 \\
75.2	0.929 \\
75.3	1.367 \\
75.4	0.896 \\
75.5	0.921 \\
75.6	1.08 \\
75.7	0.869 \\
75.8	0.951 \\
75.9	1.289 \\
76	0.966 \\
76.1	0.83 \\
76.2	0.841 \\
76.3	0.919 \\
76.4	1.069 \\
76.5	1.704 \\
76.6	0.812 \\
76.7	0.831 \\
76.8	0.919 \\
76.9	0.927 \\
77	0.999 \\
77.1	0.825 \\
77.2	0.916 \\
77.3	0.965 \\
77.4	1.118 \\
77.5	0.792 \\
77.6	0.839 \\
77.7	0.791 \\
77.8	0.892 \\
77.9	1.174 \\
78	0.8 \\
78.1	1.512 \\
78.2	0.902 \\
78.3	0.813 \\
78.4	0.925 \\
78.5	0.8 \\
78.6	0.958 \\
};
\addplot [color1]
table [row sep=\\]{%
0	54.663 \\
0.1	74.533 \\
0.2	71.331 \\
0.3	69.448 \\
0.4	72.808 \\
0.5	71.646 \\
0.6	71.25 \\
0.7	71.746 \\
0.8	71.94 \\
0.9	67.547 \\
1	75.292 \\
1.1	70.803 \\
1.2	69.744 \\
1.3	72.269 \\
1.4	68.296 \\
1.5	70.469 \\
1.6	73.564 \\
1.7	69.948 \\
1.8	69.55 \\
1.9	77.865 \\
2	71.406 \\
2.1	75.546 \\
2.2	76.599 \\
2.3	80.616 \\
2.4	73.9 \\
2.5	70.024 \\
2.6	73.532 \\
2.7	70.864 \\
2.8	73.477 \\
2.9	69.866 \\
3	72.781 \\
3.1	72.45 \\
3.2	70.226 \\
3.3	70.221 \\
3.4	71.309 \\
3.5	72.974 \\
3.6	68.539 \\
3.7	76.187 \\
3.8	75.775 \\
3.9	70.74 \\
4	70.342 \\
4.1	71.217 \\
4.2	73.899 \\
4.3	71.2 \\
4.4	68.325 \\
4.5	69.879 \\
4.6	70.968 \\
4.7	70.745 \\
4.8	70.397 \\
4.9	67.952 \\
5	70.844 \\
5.1	68.919 \\
5.2	72.157 \\
5.3	67.923 \\
5.4	68.505 \\
5.5	69.738 \\
5.6	69.566 \\
5.7	69.648 \\
5.8	68.836 \\
5.9	67.857 \\
6	68.06 \\
6.1	68.725 \\
6.2	70.635 \\
6.3	71.236 \\
6.4	69.46 \\
6.5	66.951 \\
6.6	66.619 \\
6.7	67.989 \\
6.8	69.667 \\
6.9	68.691 \\
7	75.194 \\
7.1	69.185 \\
7.2	68.131 \\
7.3	70.835 \\
7.4	69.937 \\
7.5	68.51 \\
7.6	68.338 \\
7.7	69.358 \\
7.8	68.227 \\
7.9	70.614 \\
8	68.754 \\
8.1	68.003 \\
8.2	68.599 \\
8.3	67.432 \\
8.4	68.706 \\
8.5	70.279 \\
8.6	68.093 \\
8.7	67.52 \\
8.8	68.464 \\
8.9	70.934 \\
9	72.107 \\
9.1	75.555 \\
9.2	71.782 \\
9.3	69.984 \\
9.4	73.866 \\
9.5	71.768 \\
9.6	73.269 \\
9.7	70.331 \\
9.8	70.639 \\
9.9	74.373 \\
10	70.997 \\
10.1	69.37 \\
10.2	68.646 \\
10.3	70.953 \\
10.4	71.091 \\
10.5	73.759 \\
10.6	69.452 \\
10.7	70.565 \\
10.8	73.338 \\
10.9	71.114 \\
11	76.402 \\
11.1	69.699 \\
11.2	72.464 \\
11.3	73.314 \\
11.4	70.927 \\
11.5	71.625 \\
11.6	74.709 \\
11.7	71.817 \\
11.8	70.497 \\
11.9	73.381 \\
12	70.451 \\
12.1	73.551 \\
12.2	71.06 \\
12.3	73.909 \\
12.4	74.776 \\
12.5	73.606 \\
12.6	77.665 \\
12.7	75.447 \\
12.8	78.864 \\
12.9	76.67 \\
13	76.677 \\
13.1	76.432 \\
13.2	77.052 \\
13.3	76.709 \\
13.4	74.84 \\
13.5	77.88 \\
13.6	75.559 \\
13.7	76.374 \\
13.8	74.847 \\
13.9	72.329 \\
14	72.799 \\
14.1	71.619 \\
14.2	72.382 \\
14.3	72.642 \\
14.4	70.48 \\
14.5	76.543 \\
14.6	73.336 \\
14.7	73.06 \\
14.8	76.475 \\
14.9	74.776 \\
15	71.753 \\
15.1	70.947 \\
15.2	69.216 \\
15.3	72.825 \\
15.4	72.295 \\
15.5	71.642 \\
15.6	69.735 \\
15.7	73.11 \\
15.8	73.555 \\
15.9	72.64 \\
16	72.493 \\
16.1	69.952 \\
16.2	73.046 \\
16.3	70.815 \\
16.4	71.531 \\
16.5	69.923 \\
16.6	70.057 \\
16.7	75.602 \\
16.8	72.904 \\
16.9	72.491 \\
17	74.675 \\
17.1	73.094 \\
17.2	77.722 \\
17.3	74.982 \\
17.4	78.081 \\
17.5	73.413 \\
17.6	77.265 \\
17.7	76.022 \\
17.8	77.459 \\
17.9	73.528 \\
18	75 \\
18.1	76.868 \\
18.2	78.4 \\
18.3	74.575 \\
18.4	75.988 \\
18.5	72.994 \\
18.6	76.514 \\
18.7	74.156 \\
18.8	71.615 \\
18.9	78.082 \\
19	75.316 \\
19.1	71.545 \\
19.2	73.855 \\
19.3	75.069 \\
19.4	70.917 \\
19.5	73.114 \\
19.6	74.572 \\
19.7	69.098 \\
19.8	77.785 \\
19.9	76.695 \\
20	75.812 \\
20.1	72.951 \\
20.2	75.974 \\
20.3	73.215 \\
20.4	74.14 \\
20.5	70.466 \\
20.6	70.395 \\
20.7	69.933 \\
20.8	74.758 \\
20.9	72.521 \\
21	69.821 \\
21.1	72.195 \\
21.2	72.374 \\
21.3	75.331 \\
21.4	70.745 \\
21.5	71.087 \\
21.6	73.265 \\
21.7	73.35 \\
21.8	69.729 \\
21.9	69.169 \\
22	69.717 \\
22.1	72.557 \\
22.2	70.564 \\
22.3	68.668 \\
22.4	69.5 \\
22.5	74.358 \\
22.6	78.236 \\
22.7	71.229 \\
22.8	70.141 \\
22.9	71.493 \\
23	77.493 \\
23.1	68.429 \\
23.2	68.557 \\
23.3	72.904 \\
23.4	70.034 \\
23.5	70.051 \\
23.6	70.337 \\
23.7	70.316 \\
23.8	72.132 \\
23.9	71.038 \\
24	70.248 \\
24.1	71.57 \\
24.2	73.916 \\
24.3	70.339 \\
24.4	68.583 \\
24.5	72.293 \\
24.6	70.332 \\
24.7	74.578 \\
24.8	68.754 \\
24.9	69.302 \\
25	71.262 \\
25.1	74.961 \\
25.2	71.268 \\
25.3	72.32 \\
25.4	68.77 \\
25.5	72.663 \\
25.6	71.951 \\
25.7	70.029 \\
25.8	70.524 \\
25.9	76.167 \\
26	71.939 \\
26.1	69.821 \\
26.2	76.414 \\
26.3	72.471 \\
26.4	70.801 \\
26.5	75.168 \\
26.6	73.099 \\
26.7	72.643 \\
26.8	72.582 \\
26.9	71.421 \\
27	68.758 \\
27.1	73.098 \\
27.2	74.495 \\
27.3	70.122 \\
27.4	70.314 \\
27.5	75.009 \\
27.6	72.144 \\
27.7	71.773 \\
27.8	72.785 \\
27.9	70.772 \\
28	69.346 \\
28.1	71.493 \\
28.2	70.049 \\
28.3	70.223 \\
28.4	72.702 \\
28.5	70.963 \\
28.6	72.236 \\
28.7	76.654 \\
28.8	74.771 \\
28.9	70.259 \\
29	68.911 \\
29.1	70.84 \\
29.2	74.784 \\
29.3	72.516 \\
29.4	72.988 \\
29.5	73.262 \\
29.6	70.175 \\
29.7	80.306 \\
29.8	69.798 \\
29.9	70.007 \\
30	75.923 \\
30.1	71.961 \\
30.2	70.769 \\
30.3	73.361 \\
30.4	72.092 \\
30.5	66.811 \\
30.6	70.452 \\
30.7	71.853 \\
30.8	69.676 \\
30.9	76.775 \\
31	73.226 \\
31.1	72.532 \\
31.2	72.678 \\
31.3	71.338 \\
31.4	73.919 \\
31.5	71.502 \\
31.6	71.228 \\
31.7	72.731 \\
31.8	70.469 \\
31.9	69.419 \\
32	72.879 \\
32.1	69.49 \\
32.2	67.931 \\
32.3	67.405 \\
32.4	72.252 \\
32.5	69.427 \\
32.6	69.967 \\
32.7	69.697 \\
32.8	68.173 \\
32.9	68.971 \\
33	70.606 \\
33.1	66.766 \\
33.2	68.309 \\
33.3	67.177 \\
33.4	71.96 \\
33.5	72.468 \\
33.6	69.829 \\
33.7	69.483 \\
33.8	67.554 \\
33.9	68.464 \\
34	69.843 \\
34.1	69.343 \\
34.2	70.46 \\
34.3	67.976 \\
34.4	69.074 \\
34.5	70.427 \\
34.6	72.402 \\
34.7	75.858 \\
34.8	69.304 \\
34.9	73.44 \\
35	72.945 \\
35.1	73.564 \\
35.2	74.213 \\
35.3	68.973 \\
35.4	70.401 \\
35.5	71.708 \\
35.6	69.73 \\
35.7	70.397 \\
35.8	68.246 \\
35.9	72.244 \\
36	71.017 \\
36.1	70.45 \\
36.2	69.104 \\
36.3	69.547 \\
36.4	69.961 \\
36.5	71.548 \\
36.6	74.409 \\
36.7	72.229 \\
36.8	74.976 \\
36.9	75.103 \\
37	73.138 \\
37.1	73.369 \\
37.2	74.683 \\
37.3	85.376 \\
37.4	75.768 \\
37.5	77.135 \\
37.6	75.546 \\
37.7	79.342 \\
37.8	78.991 \\
37.9	80.186 \\
38	79.581 \\
38.1	80.55 \\
38.2	79.482 \\
38.3	80.232 \\
38.4	75.68 \\
38.5	79.244 \\
38.6	75.532 \\
38.7	78.883 \\
38.8	76.776 \\
38.9	76.971 \\
39	74.237 \\
39.1	76.891 \\
39.2	73.689 \\
39.3	77.473 \\
39.4	72.345 \\
39.5	75.666 \\
39.6	74.109 \\
39.7	73.963 \\
39.8	73.627 \\
39.9	70.881 \\
40	72.504 \\
40.1	70.448 \\
40.2	71.363 \\
40.3	74.011 \\
40.4	70.552 \\
40.5	70.793 \\
40.6	71.807 \\
40.7	70.04 \\
40.8	70.51 \\
40.9	72.84 \\
41	71.45 \\
41.1	70.505 \\
41.2	70.892 \\
41.3	71.357 \\
41.4	71.875 \\
41.5	70.34 \\
41.6	71.289 \\
41.7	75.44 \\
41.8	72.269 \\
41.9	70.657 \\
42	73.42 \\
42.1	72.013 \\
42.2	76.665 \\
42.3	70.395 \\
42.4	71.263 \\
42.5	75.023 \\
42.6	72.602 \\
42.7	77.292 \\
42.8	70.406 \\
42.9	70.025 \\
43	72.403 \\
43.1	69.57 \\
43.2	73.002 \\
43.3	72.119 \\
43.4	69.721 \\
43.5	70.528 \\
43.6	70.604 \\
43.7	69.597 \\
43.8	69.261 \\
43.9	70.765 \\
44	70.465 \\
44.1	70.943 \\
44.2	71.274 \\
44.3	72.728 \\
44.4	69.651 \\
44.5	69.383 \\
44.6	71.601 \\
44.7	69.538 \\
44.8	68.339 \\
44.9	71.357 \\
45	71.267 \\
45.1	72.374 \\
45.2	72.05 \\
45.3	71.694 \\
45.4	76.712 \\
45.5	69.961 \\
45.6	72.89 \\
45.7	75.276 \\
45.8	75.788 \\
45.9	77.803 \\
46	73.628 \\
46.1	83.799 \\
46.2	76.719 \\
46.3	74.215 \\
46.4	76.933 \\
46.5	74.711 \\
46.6	74.136 \\
46.7	72.16 \\
46.8	75.328 \\
46.9	76.279 \\
47	70.947 \\
47.1	74.997 \\
47.2	71.741 \\
47.3	71.053 \\
47.4	72.336 \\
47.5	71.713 \\
47.6	73.155 \\
47.7	71.189 \\
47.8	70.206 \\
47.9	72.028 \\
48	71.91 \\
48.1	74.771 \\
48.2	76.124 \\
48.3	73.886 \\
48.4	74.151 \\
48.5	71.589 \\
48.6	74.187 \\
48.7	74.827 \\
48.8	72.82 \\
48.9	73.861 \\
49	73.174 \\
49.1	75.508 \\
49.2	80.977 \\
49.3	91.509 \\
49.4	74.689 \\
49.5	71.478 \\
49.6	71.351 \\
49.7	75.858 \\
49.8	73.901 \\
49.9	81.798 \\
50	75.66 \\
50.1	71.57 \\
50.2	76.88 \\
50.3	72.625 \\
50.4	72.697 \\
50.5	79.654 \\
50.6	77.04 \\
50.7	77.692 \\
50.8	77.802 \\
50.9	73.755 \\
51	82.823 \\
51.1	78.583 \\
51.2	82.442 \\
51.3	78.538 \\
51.4	79.867 \\
51.5	85.652 \\
51.6	81.399 \\
51.7	77.094 \\
51.8	80.699 \\
51.9	76.295 \\
52	78.034 \\
52.1	76.929 \\
52.2	84.286 \\
52.3	86.951 \\
52.4	87.577 \\
52.5	80.115 \\
52.6	80.272 \\
52.7	83.688 \\
52.8	83.658 \\
52.9	83.033 \\
53	80.007 \\
53.1	75.378 \\
53.2	76.378 \\
53.3	76.169 \\
53.4	75.216 \\
53.5	78.341 \\
53.6	80.395 \\
53.7	81.887 \\
53.8	79.992 \\
53.9	77.331 \\
54	75.103 \\
54.1	72.863 \\
54.2	69.914 \\
54.3	73.533 \\
54.4	72.029 \\
54.5	76.566 \\
54.6	72.977 \\
54.7	73.611 \\
54.8	75.765 \\
54.9	78.791 \\
55	74.629 \\
55.1	70.683 \\
55.2	76.155 \\
55.3	83.048 \\
55.4	77.086 \\
55.5	78.259 \\
55.6	76.659 \\
55.7	78.45 \\
55.8	76.595 \\
55.9	77.395 \\
56	76.407 \\
56.1	72.474 \\
56.2	72.994 \\
56.3	72.163 \\
56.4	75.948 \\
56.5	70.124 \\
56.6	78.183 \\
56.7	76.329 \\
56.8	75.53 \\
56.9	76.406 \\
57	77.805 \\
57.1	75.251 \\
57.2	75.293 \\
57.3	76.596 \\
57.4	79.484 \\
57.5	73.894 \\
57.6	73.991 \\
57.7	72.543 \\
57.8	73.236 \\
57.9	71.905 \\
58	70.719 \\
58.1	75.75 \\
58.2	71.943 \\
58.3	71.985 \\
58.4	72.219 \\
58.5	74.463 \\
58.6	76.009 \\
58.7	71.875 \\
58.8	71.101 \\
58.9	72.964 \\
59	72.967 \\
59.1	73.6 \\
59.2	70.393 \\
59.3	71.113 \\
59.4	74.193 \\
59.5	72.232 \\
59.6	75.239 \\
59.7	71.905 \\
59.8	73.136 \\
59.9	74.924 \\
60	72.032 \\
60.1	73.78 \\
60.2	76.638 \\
60.3	78.721 \\
60.4	92.555 \\
60.5	71.231 \\
60.6	74.88 \\
60.7	72.517 \\
60.8	70.084 \\
60.9	81.37 \\
61	74.828 \\
61.1	76.199 \\
61.2	72.314 \\
61.3	75.919 \\
61.4	78.388 \\
61.5	71.984 \\
61.6	75.617 \\
61.7	75.335 \\
61.8	73.665 \\
61.9	69.459 \\
62	72.273 \\
62.1	72.085 \\
62.2	73.578 \\
62.3	74.135 \\
62.4	72.169 \\
62.5	78.145 \\
62.6	72.708 \\
62.7	78.625 \\
62.8	70.782 \\
62.9	77.028 \\
63	74.009 \\
63.1	78.487 \\
63.2	75.487 \\
63.3	76.617 \\
63.4	75.364 \\
63.5	80.221 \\
63.6	81.34 \\
63.7	77.805 \\
63.8	84.086 \\
63.9	75.372 \\
64	81.445 \\
64.1	82.257 \\
64.2	79.347 \\
64.3	75.793 \\
64.4	78.987 \\
64.5	81.29 \\
64.6	80.787 \\
64.7	76.748 \\
64.8	78.103 \\
64.9	77.091 \\
65	75.836 \\
65.1	78.482 \\
65.2	75.026 \\
65.3	76.023 \\
65.4	74.006 \\
65.5	76.068 \\
65.6	72.936 \\
65.7	72.34 \\
65.8	73.137 \\
65.9	71.489 \\
66	78.392 \\
66.1	74.703 \\
66.2	80.459 \\
66.3	75.353 \\
66.4	74.113 \\
66.5	72.362 \\
66.6	72.991 \\
66.7	71.471 \\
66.8	73.394 \\
66.9	76.577 \\
67	71.333 \\
67.1	77.244 \\
67.2	72.359 \\
67.3	72.917 \\
67.4	71.656 \\
67.5	77.121 \\
67.6	72.292 \\
67.7	71.105 \\
67.8	74.37 \\
67.9	93.326 \\
68	74.011 \\
68.1	77.802 \\
68.2	70.567 \\
68.3	71.561 \\
68.4	73.666 \\
68.5	74.209 \\
68.6	73.263 \\
68.7	71.929 \\
68.8	71.534 \\
68.9	74.892 \\
69	71.557 \\
69.1	76.705 \\
69.2	75.873 \\
69.3	76.885 \\
69.4	72.828 \\
69.5	74.247 \\
69.6	72.915 \\
69.7	71.762 \\
69.8	75.014 \\
69.9	77.541 \\
70	76.973 \\
70.1	76.751 \\
70.2	78.323 \\
70.3	72.463 \\
70.4	72.449 \\
70.5	72.086 \\
70.6	73.69 \\
70.7	76.086 \\
70.8	76.505 \\
70.9	75.961 \\
71	77.788 \\
71.1	77.581 \\
71.2	79.252 \\
71.3	79.95 \\
71.4	75.772 \\
71.5	74.803 \\
71.6	73.431 \\
71.7	72.106 \\
71.8	70.888 \\
71.9	75.456 \\
72	73.802 \\
72.1	76.588 \\
72.2	76.102 \\
72.3	74.7 \\
72.4	74.13 \\
72.5	74.858 \\
72.6	75.253 \\
72.7	76.93 \\
72.8	72.372 \\
72.9	72.158 \\
73	76.12 \\
73.1	72.047 \\
73.2	71.32 \\
73.3	73.283 \\
73.4	73.464 \\
73.5	73.687 \\
73.6	70.434 \\
73.7	71.536 \\
73.8	76.661 \\
73.9	73.626 \\
74	73.916 \\
74.1	72.788 \\
74.2	75.315 \\
74.3	72.705 \\
74.4	70.477 \\
74.5	68.786 \\
74.6	71.974 \\
74.7	71.724 \\
74.8	70.968 \\
74.9	70.496 \\
75	72.498 \\
75.1	74.738 \\
75.2	74.077 \\
75.3	75.596 \\
75.4	72.821 \\
75.5	72.122 \\
75.6	71.12 \\
75.7	71.639 \\
75.8	71.103 \\
75.9	76.023 \\
76	74.573 \\
76.1	74.199 \\
76.2	71.957 \\
76.3	71.283 \\
76.4	81.182 \\
76.5	75.336 \\
76.6	78.49 \\
76.7	71.484 \\
76.8	74.013 \\
76.9	74.475 \\
77	79.661 \\
77.1	77.804 \\
77.2	82.218 \\
77.3	72.755 \\
77.4	75.536 \\
77.5	75.885 \\
77.6	78.692 \\
77.7	77.118 \\
77.8	79.307 \\
77.9	75.694 \\
78	77.84 \\
78.1	78.092 \\
78.2	74.238 \\
78.3	72.053 \\
78.4	75.729 \\
78.5	74.212 \\
78.6	79.379 \\
};
\end{axis}

\end{tikzpicture}